\documentclass{article} 
\usepackage{iclr2025_conference,times}
\iclrfinalcopy

\usepackage{amsmath,amsfonts,bm}

\def\eqref#1{equation~\ref{#1}}

\def\1{\bm{1}}

\DeclareMathAlphabet{\mathsfit}{\encodingdefault}{\sfdefault}{m}{sl}
\SetMathAlphabet{\mathsfit}{bold}{\encodingdefault}{\sfdefault}{bx}{n}

\usepackage{subcaption}
\usepackage{amsmath}
\usepackage{amssymb}
\usepackage{graphicx}
\usepackage{booktabs}
\usepackage[hyphens]{url}
\usepackage{hyperref}
\usepackage{lipsum} 
\usepackage{enumitem}
\usepackage{fancyvrb}
\usepackage[many]{tcolorbox}
\usepackage{verbatim}
\hypersetup{
    colorlinks=true,
    linkcolor=blue,
    filecolor=magenta,      
    urlcolor=cyan,
    citecolor=blue,
}
\usepackage{multirow}
\usepackage[ruled,vlined]{algorithm2e}
\newcommand{\alias}{VIPER-R1}

\title{Mimicking the Physicist's Eye : A VLM-centric Approach for Physics Formula Discovery}

\newcommand{\namesep}{\hspace{0.55em}}   
\newcommand{\affsep}{\hspace{0.75em}}    

\author{%
\parbox{\linewidth}{\centering
\vspace{1em}
\textbf{Jiaqi Liu}\textsuperscript{1,3}\namesep
\textbf{Songning Lai}\textsuperscript{2}\namesep
\textbf{Pengze Li}\textsuperscript{3,4}\namesep
\textbf{Di Yu}\textsuperscript{3,5}\namesep
\textbf{Wenjie Zhou}\textsuperscript{6,9}\\[0.3em]
\textbf{Yiyang Zhou}\textsuperscript{1}\namesep
\textbf{Peng Xia}\textsuperscript{1}\namesep
\textbf{Zijun Wang}\textsuperscript{7}\namesep
\textbf{Xi Chen}\textsuperscript{4}\namesep
\textbf{Shixiang Tang}\textsuperscript{8}\\[0.3em]
\textbf{Lei Bai}\textsuperscript{3}\namesep
\textbf{Wanli Ouyang}\textsuperscript{3,8}\namesep
\textbf{Mingyu Ding}\textsuperscript{1}\namesep
\textbf{Huaxiu Yao}\textsuperscript{1}\namesep
\textbf{Aoran Wang}\textsuperscript{3}\thanks{Corresponding author}
\\[0.45em]
{\normalfont\small
$^1$ UNC--Chapel Hill\affsep
$^2$ HKUST (Guangzhou)\affsep
$^3$ Shanghai Artificial Intelligence Laboratory\\[0.35em]
$^4$ Fudan University\affsep
$^5$ Tsinghua University\affsep
$^6$ Nankai University\affsep
$^7$ UC Santa Cruz\\[0.35em]
$^8$ The Chinese University of Hong Kong\affsep
$^9$ Shanghai Innovation Institute\\[0.25em]
{\ttfamily\mdseries \{jqliu@cs.unc.edu,\enspace\; wangaoran@pjlab.org.cn}\}
}
}
}

\begin{document}

\maketitle
\begin{figure*}[ht]
  \centering
  \vspace{-0.5cm}
  \includegraphics[width=\linewidth]{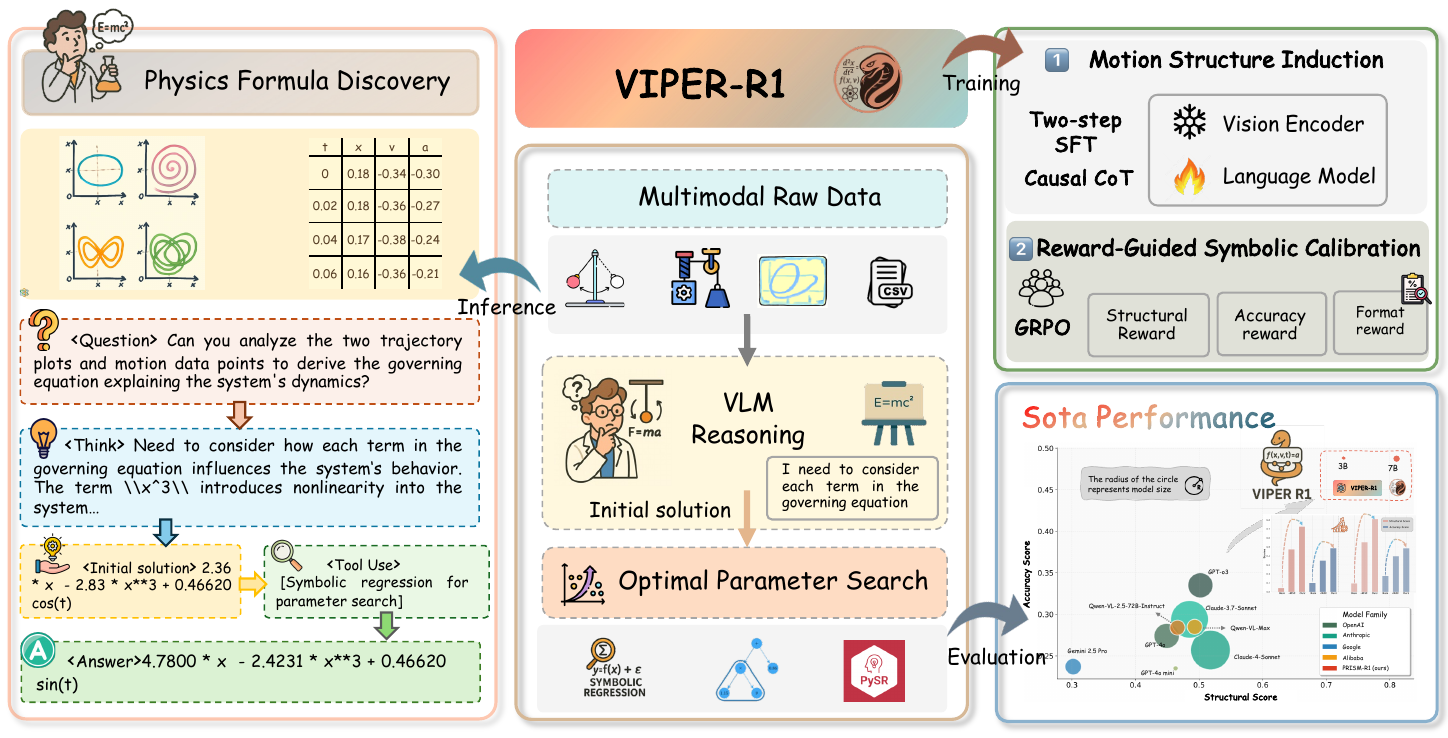}
  \caption{
  Overview of \textbf{VIPER-R1}, a multimodal framework for physics formula discovery. 
The model is trained via Motion Structure Induction (MSI) with Causal CoT supervision and Reward-Guided Symbolic Calibration (RGSC) for structural refinement. 
During inference, VIPER-R1 acts agentically by invoking an external symbolic regression tool for Symbolic Residual Realignment (SR²), reconciling symbolic hypotheses with empirical data. The model achieves state-of-the-art performance in both structural and accuracy scores on the PhysSymbol dataset.
  }
  \label{fig:overview}
  \vspace{4pt}
\end{figure*}

\begin{abstract}
Automated discovery of physical laws from observational data in the real world is a grand challenge in AI. Current methods, relying on symbolic regression or LLMs, are limited to uni-modal data and overlook the rich, visual phenomenological representations of motion that are indispensable to physicists. This ``sensory deprivation" severely weakens their ability to interpret the inherent spatio-temporal patterns within dynamic phenomena.
To address this gap, we propose \textbf{VIPER-R1}, a multimodal model that performs \textbf{V}isual \textbf{I}nduction for \textbf{P}hysics-based \textbf{E}quation \textbf{R}easoning to discover fundamental symbolic formulas. It methodically integrates visual perception, trajectory data, and symbolic reasoning to simulate the scientific discovery process.
The model is trained via a curriculum of Motion Structure Induction (MSI), using supervised fine-tuning to interpret kinematic phase portraits and construct hypotheses guided by a Causal Chain of Thought (C-CoT), followed by Reward-Guided Symbolic Calibration (RGSC) to purify the formula's structure with reinforcement learning. 
During inference, the trained \textbf{VIPER-R1} acts as an agent: it first posits a high-confidence symbolic ansatz, then proactively invokes an external symbolic regression tool to perform Symbolic Residual Realignment (SR²). This final step, analogous to a physicist’s perturbation analysis, reconciles the theoretical model with empirical data.
To support this research, we introduce PhysSymbol, a new 5,000-instance multimodal corpus. Experiments show that \textbf{VIPER-R1} consistently outperforms state-of-the-art VLM baselines in accuracy and interpretability, enabling more precise discovery of physical laws\footnote[1]{Project in: \url{https://jiaaqiliu.github.io/VIPER-R1/}}.
\end{abstract}

\section{Introduction}
The automated discovery of fundamental physical laws in the form of equations from observational data stands as a grand challenge at the intersection of artificial intelligence and the natural sciences~\citep{Udrescu2020AI, Wang2023scientific-age-AI}. This endeavor is pivotal for augmenting human scientific intuition and accelerating the pace of discovery by uncovering novel principles within vast, high-dimensional datasets~\citep{lu2024ai, reddy2024towards}. Recent advances have established two parallel yet distinct research tracks: sophisticated symbolic regression (SR) algorithms that navigate immense combinatorial spaces to identify fitting equations~\citep{SRBench-Cava-NeurIPS-2021, Cranmer2023Interpretable}, and the emergence of Large Language Models (LLMs) demonstrating remarkable ability to perform in-context symbolic reasoning from textual data~\citep{ma2024llm, grayeli2024symbolic, Shojaee2025LLMSR}. While both approaches have laid critical foundations, they share a  disconnect with the actual process of human scientific inquiry, operating without a key perceptual faculty that is central to human discovery.

This limitation can be seen as a form of ``sensory deprivation,'' where reliance on uni-modal symbolic data blinds models to the rich visual representations that physicists routinely exploit. 
Human scientific reasoning is inherently multimodal: physicists interpret visual patterns in phase portraits to infer conservation laws, recognize decay envelopes to hypothesize damping forces, and identify superposition effects to constrain theoretical possibilities~\citep{strogatz2018nonlinear}. 
Such visual intuition provides powerful pre-symbolic heuristics for navigating the vast space of candidate theories.

Recent advances in LLM-based scientific discovery partly address these issues. 
LLM-SR~\citep{Shojaee2025LLMSR} generates equation hypotheses from embedded scientific knowledge, while frameworks like Scientific Generative Agents~\citep{ma2024llm} pair LLM-based generation with simulation validation. 
Yet these methods still suffer from ``sensory deprivation,'' lacking the ability to incorporate visual evidence. 
Furthermore, concerns about memorization versus genuine discovery~\citep{wu2023reasoning, Shojaee2025LLMSRBench} underscore the need for approaches that perform authentic data-driven reasoning rather than recalling known formulas.

By neglecting the crucial visual perceptual channel, existing methods are fundamentally constrained. They often resort to computationally expensive searches through vast equation spaces~\citep{Virgolin2022NP-hard}, exhibit brittle token-matching behaviors, and fail to achieve the intuitive leaps that characterize human scientific breakthroughs. This limitation becomes particularly pronounced when dealing with complex dynamical systems where visual patterns in phase space and temporal evolution provide crucial insights that are difficult to extract from purely numerical data.

To bridge the gap between raw perception and abstract formalism, we introduce \textbf{VIPER-R1}, a \textbf{V}isual \textbf{I}nduction model for \textbf{P}hysics-based \textbf{E}quation \textbf{R}easoning. 
Rather than a mere pattern matcher, VIPER-R1 acts as a ``computational phenomenologist," grounding symbolic reasoning in visual evidence by integrating plots, trajectory data, and symbolic logic to autonomously derive governing laws of motion.

Our framework draws inspiration from human scientific reasoning and follows a two-stage pipeline. 
In the first stage, \textbf{Motion Structure Induction (MSI)}, the model undergoes Supervised Fine-Tuning (SFT), learning to interpret kinematic evidence under joint supervision of Chain-of-Thought (CoT) rationales and ground-truth equations, before producing initial symbolic hypotheses guided by causal CoT prompts. 
In the second stage, \textbf{Reward-Guided Symbolic Calibration (RGSC)}, reinforcement learning with Group Relative Policy Optimization (GRPO)~\citep{grpo} refines these hypotheses using a structural reward function that favors topological correctness over coefficient matching. 
Finally, the model invokes an external symbolic regression tool for \textbf{Symbolic Residual Realignment (SR²)}, aligning theoretical expressions with empirical details to yield interpretable, precise formulas.

To support this research, we also release \textbf{PhysSymbol}, a large-scale multimodal corpus of 5,000 instances designed for training and evaluating models on physics formula discovery.

Our contributions can be summarized as follows:
\begin{itemize}[labelsep = .5em, leftmargin = 0pt, itemindent = 1em]
    \item We propose \textbf{\alias}, a multimodal framework that simulates the scientific reasoning process by deeply integrating visual perception with symbolic derivation.
    \item A two-step training and inference strategy is designed, Motion Structure Induction (MSI) for hypothesis generation and Reward-Guided Symbolic Calibration (RGSC) for structural refinement. 
    \item We introduce an agentic refinement stage, Symbolic Residual Realignment (SR²), where the VLM proactively utilizes external tools to harmonize its theoretical hypotheses with empirical data, aligning with modern agent-based AI paradigms.
    \item We introduce \textbf{PhysSymbol}, a large-scale benchmark of 5,000 multimodal physics instances, created to advance research in vision-grounded scientific discovery.
\end{itemize}

\begin{figure}
    \centering
    \includegraphics[width=0.85\linewidth]{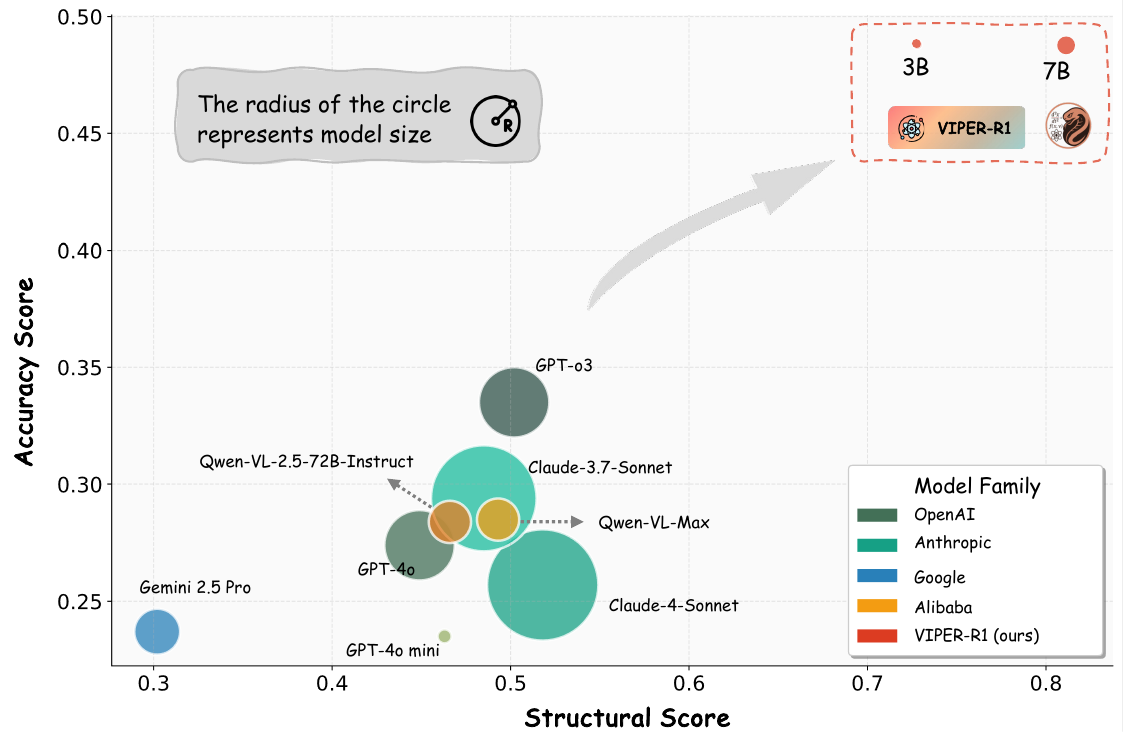}
    \caption{The performance of different SOTA VLMs on physics formula discovery tasks.}
    \label{fig:performance_bubble}
\end{figure}

\section{Related Work}

\subsection{Symbolic Regression for Scientific Discovery}
Symbolic regression (SR) aims to discover mathematical expressions from data, a field founded on techniques like genetic programming~\citep{koza1994genetic}. Modern methods have significantly advanced this area, with physics-inspired recursive algorithms like AI Feynman~\citep{Udrescu2020AI} and high-performance evolutionary tools like PySR~\citep{Cranmer2023Interpretable}. Recent deep learning approaches leverage Transformer architectures to map numerical data directly to symbolic expressions~\citep{Biggio-NeSymReS-ICML-2021, Kamienny-E2E-symbolic-NIPS-2022}, while hybrid systems combine neural networks with methods like reinforcement learning~\citep{DSR-Petersen-ICLR-2021}, Monte Carlo tree search~\citep{SPL-MCTS-ICLR-2023}, and guided genetic programming~\citep{mundhenk-seeding-GP-NeurIPS-2021, SNIP-ICLR}. Despite these advances, SR faces the persistent, NP-hard challenge~\citep{Virgolin2022NP-hard, Shojaee2025LLMSR} of navigating a vast search space without strong priors, often leading to computationally expensive searches for physically implausible equations~\citep{Virgolin2022NP-hard}. Our work confronts this by using a VLM to generate a strong, visually-grounded prior, transforming SR from a blind search into a targeted refinement.



\subsection{LLMs for Scientific Discovery}
The advent of LLMs has created transformative possibilities for automating science~\citep{Wang2023scientific-age-AI, lu2024ai}. Several recent frameworks now leverage LLMs for equation discovery by generating equation skeletons~\citep{Shojaee2025LLMSR}, using in-context learning~\citep{merler2024context}, implementing bilevel optimization with simulators~\citep{ma2024llm}, and building libraries of scientific concepts~\citep{grayeli2024symbolic}. A key concern in this area is the models' tendency to memorize formulas, an issue addressed by specialized benchmarks~\citep{wu2023reasoning, mirzadeh2024gsm, Shojaee2025LLMSRBench}. Concurrently, LLMs are being explored as powerful optimization and evolution engines~\citep{Lehman-ELM, FunSearch, LLM-Evo-Lange-2024} for tasks such as prompt optimization~\citep{connect-LLM-Evo-prompt, LLM-Optimizer}, neural architecture search~\citep{EvoPrompt-2023, gpt4-NAS-llm}, and heuristic discovery. While LLMs also demonstrate remarkable capabilities in general scientific hypothesis generation and reasoning~\citep{LLM-scientific-synthesis, LLM-scientific-zero-shot-hypothesis, wang2023hypothesis, data-driven-AI2-LM, li2024automated, ji2024scimon, ma2024llm-bilevel}, their uni-modal nature renders them blind to the holistic visual patterns apparent to human scientists. Our work bridges this sensory gap.



\subsection{Multimodal Models for Scientific Discovery}
 VLMs are increasingly being applied in scientific domains for their ability to reason about visual content~\citep{zhang2024comprehensive,su2025thinking}, from interpreting research figures~\citep{Lu2022Learna, zhang2024vision} to general scientific understanding with models like GPT-4V~\citep{openai2024gpt4ocard}, Qwen-VL~\citep{Bai2023QwenVL}, and Gemini~\citep{gemini25}. Pioneering work has utilized Multimodal LLMs to discover governing equations from video data by first identifying intrinsic coordinates and then performing symbolic reasoning~\citep{Li2025MLLMbased}. Our work addresses a different and complementary aspect of the scientific workflow; we focus on the 2D graphical representations (e.g., phase plots) that scientists create for analysis. The VLM's role is not coordinate discovery but direct visual reasoning on these plots to hypothesize functional forms, mimicking a physicist who recognizes patterns like ``damped oscillation" and sketches initial formulas. While many scientific benchmarks exist~\citep{SRBench-Cava-NeurIPS-2021, rethinking-scientific-2022, majumder2024discoverybench, wang2025physunibench}, they often face memorization issues with LLMs~\citep{Shojaee2025LLMSRBench}. Our approach is the first to leverage a fine-tuned VLM for direct, plot-based hypothesis generation in physics, more closely emulating the human observation-and-reasoning cycle.

\begin{figure*}[t]
  \centering
  \includegraphics[width=\linewidth]{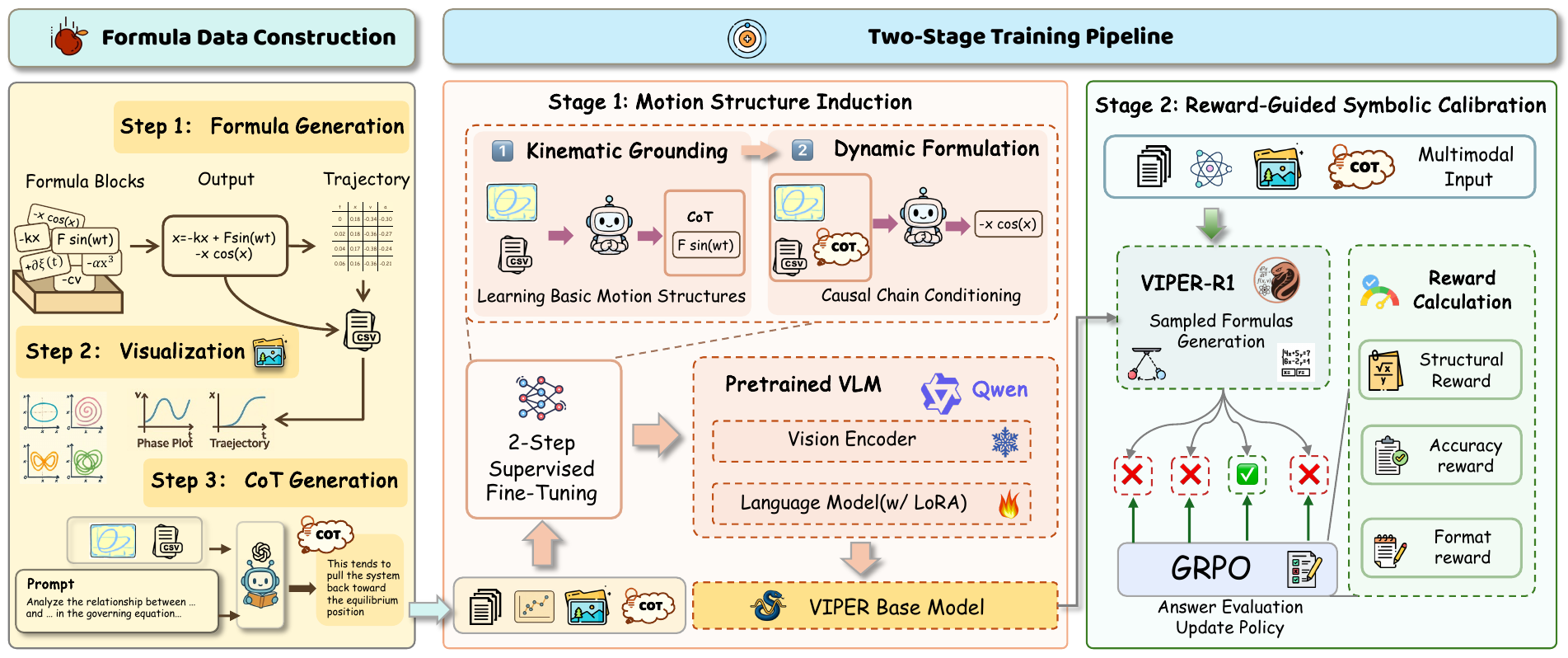}
  \caption{\textbf{Framework of VIPER-R1}.
   VIPER-R1 introduces a two-phase training framework for visual formula discovery and reasoning. First, a two-step curriculum, called  Motion Structure Induction (MSI), is designed to imbue the VIPER-R1 with the ability to deduce the latent symbolic structure of a system's dynamics at stage 1. Subsequently, in stage 2, we employ reinforcement learning to ``anneal" the model's generation policy, sharpening its focus on producing topologically correct physical laws. }
  \label{fig:framework}
  \vspace{-10pt}
\end{figure*}

\section{Methodology}

Our proposed framework consists of a two-stage pipeline, as illustrated in Figure \ref{fig:framework}. The first stage involves two-step Motion Structure Induction with CoT reasoning that activates the model's reasoning potential and the ability of formula structure induction. At stage 2, a RL-based refinement method is employed to help the model further calibrate the symbolic solution.
When inferencing, a symbolic regression module is design as a optimal parameter searching tool to refine this hypothesis.

\subsection{Problem Definition}
The automated discovery of physical laws from multimodal empirical data can be formally defined as learning a mapping from a set of observations to the underlying symbolic law that governs the system. This process seeks to infer an interpretable symbolic expression $S$ from a diverse set of empirical evidence $\mathcal{E}$. The mapping can be represented as:
\[
\pi_\theta : \mathcal{E} \rightarrow E,
\]
where:
\begin{itemize}
    \item $\mathcal{E} = \{\mathcal{V}, \mathcal{D}\}$ represents the complete set of Empirical Evidence, comprising both visual and numerical data modalities.
    
    \item $\mathcal{V} = \{V_1, V_2, \dots\}$ is a set of visual representations of the system's dynamics. For instance, in the context of the kinematic systems studied in this work, $\mathcal{V}$ typically includes a phase-space portrait ($I_{\text{phase}}$) and a time-series trajectory plot ($I_{\text{trajectory}}$). More broadly, $\mathcal{V}$ could encompass video frames of a real-world experiment, heatmaps of a field distribution, or other scientific visualizations.
    
    \item $\mathcal{D} = \{D_1, D_2, \dots\}$ is a set of quantitative measurements of the system's state variables. For the mechanical systems we investigate, $\mathcal{D}$ consists of time-series data of position, velocity, and acceleration, i.e., $\{ (t_i, x(t_i), v(t_i), a(t_i)) \}$.
    
    \item $E$ is the target output: an interpretable symbolic expression representing the governing physical law.
    
    \item $\pi_\theta$ is the parameterized model (in our case, the VIPER-R1) that we aim to train. 
\end{itemize}

Through this mapping, our system integrates both visual information from dynamic plots and structured motion data to emulate the observation‑and‑reasoning workflow of physicists.

\subsection{Motion Structure Induction (MSI)}

The foundational stage of our framework is Motion Structure Induction (MSI), a specialized two-step curriculum designed to imbue the VIPER-R1 with the ability to deduce the latent symbolic structure of a system's dynamics from its visual phenomenological representations. This process explicitly emulates the cognitive progression from qualitative observation to quantitative hypothesis.

\subsubsection{Step 1: Joint Induction of Causal Reasoning and Symbolic Structure}
The initial stage mirrors a physicist's first encounter with a new phenomenon: concurrently observing, reasoning, and formulating a preliminary idea. Here, the VIPER-R1 is trained to jointly generate both a Causal Chain of Thought (C-CoT) and an initial Symbolic Ansatz. The input is the complete set of Empirical Evidence $E=(\mathcal{V}, \mathcal{D})$. The model's objective is to maximize the likelihood of the entire structured output, which comprises the reasoning chain $C$ followed by the symbolic law $S$.

This joint objective is crucial; it compels the VIPER-R1 to ground its symbolic output in an explicit, physically-motivated reasoning process. The model must learn not just \textit{what} the governing law is, but \textit{why} it takes that form, based on visual cues within the evidence. Formally, we define the training objective for this stage by maximizing the log-probability of the target sequence $Y=(C,S)$:
\begin{equation}
    \mathcal{L}_{\text{MSI-1}} = -\mathbb{E}_{(E, Y) \sim \mathcal{D}_{\text{phys}}} \sum_{t=1}^{|Y|} \log \pi_{\theta}(y_t \mid E, y_{<t}),
    \label{eq:msi1}
\end{equation}
where $\mathcal{D}$ is our PhysSymbol Corpus, $Y = (y_1, ..., y_{|Y|})$ is the concatenated sequence of the C-CoT and the symbolic law, and $\pi_{\theta}$ is the policy of the VIPER-R1.

\subsubsection{Step 2: C-CoT-Guided Symbolic Formulation}
The second stage of our curriculum refines the VIPER-R1's ability to translate a well-formed physical argument into a precise symbolic form. This is analogous to a physicist taking their detailed notes and meticulously composing the final equation. In this stage, the model is provided with both the empirical evidence $E$ and the ground-truth C-CoT, $C$, and is tasked \textit{only} with generating the correct symbolic law $S$.

By conditioning on an ideal reasoning chain, we allow the model to dedicate its full representational capacity to mastering the complex syntax and semantics of physical formalisms. This decouples the task of reasoning from the task of formulation. The loss is computed exclusively on the tokens of the symbolic law $S$:
\begin{equation}
    \mathcal{L}_{\text{MSI-2}} = -\mathbb{E}_{(E, C, S) \sim \mathcal{D}_{\text{phys}}} \sum_{t=1}^{|S|} \log \pi_{\theta}(s_t \mid E, C, s_{<t}).
    \label{eq:msi2}
\end{equation}

The two‑stage MSI curriculum is designed with two key considerations. First, by decoupling the complex cognitive task of causal reasoning from the intricate syntactic task of symbolic formulation, it enhances both the stability and the effectiveness of learning. Second, this curriculum reflects a hierarchical abstraction process: it encourages the model to first construct a high‑level qualitative understanding through C‑CoT, and only then proceed to generate a low‑level, precise symbolic output—thereby mirroring effective human problem‑solving strategies.

Upon completion of MSI, the resulting model, $\pi_{\text{VIPER}}$, possesses a robust, physically-grounded foundation, ready for the subsequent Reward-Guided Symbolic Calibration stage.

\subsection{Reward-Guided Symbolic Calibration (RGSC)}

Following the foundational MSI phase, the VIPER-R1 possesses the ability to generate plausible symbolic hypotheses. However, to further enhance the structural purity and reliability of these hypotheses, we introduce a refinement phase: Reward-Guided Symbolic Calibration (RGSC). This stage employs reinforcement learning to ``anneal" the model's generation policy, sharpening its focus on producing topologically correct physical laws. We select the Group Relative Policy Optimization (GRPO) algorithm \citep{grpo} for this task, as it is highly efficient for large-scale models and circumvents the need for a separate, computationally expensive value network. GRPO's design, which computes advantages relative to a batch of sampled actions, is exceptionally well-suited for our task where a direct, analytical reward can be computed for any generated symbolic expression.

\vspace{\baselineskip}
\noindent\textbf{Sampling a Distribution of Symbolic Hypotheses.}
For each instance of Empirical Evidence $E=(\mathcal{V}, \mathcal{D})$ from our PhysSymbol Corpus, we sample a group of $G$ candidate symbolic expressions $\{S_1, S_2, \dots, S_G\}$ from the current policy $\pi_{\theta}$, which is initialized from the model fine-tuned during MSI. This sampling process is defined as:
\begin{equation}
    S_i \sim \pi_{\theta}(S \mid E), \quad \text{for } i = 1, 2, \dots, G.
\end{equation}
This strategy encourages exploration within the vast space of possible physical theories, allowing the model to discover and reinforce more robust and accurate symbolic structures.

\vspace{\baselineskip}
\noindent\textbf{Formulating the Structural Reward.}
Each sampled ansatz $S_i$ is evaluated and assigned a reward $R(S_i)$. Our reward function is designed to align with the central goal of discovering structurally correct physical laws, regardless of specific coefficient values. 
It consists of three weighted components: a Format Reward ($R_{\text{format}}$), our novel Parameter-Agnostic Structural Reward ($R_{\text{structural}}$), and an Exact Match Accuracy Reward ($R_{\text{accuracy}}$).
\begin{equation}
    R(S_i) = w_f R_{\text{format}}(S_i) + w_s R_{\text{structural}}(S_i, S_{\text{GT}}) + w_a R_{\text{accuracy}}(S_i, S_{\text{GT}}).
\end{equation}

\begin{itemize}[labelsep = .5em, leftmargin = 0pt, itemindent = 1em]
    \item \textbf{Format Reward ($R_{\text{format}}$):} This binary reward component ensures the model's output adheres strictly to the predefined \texttt{<think>...<answer>} template, which is crucial for interpretability and reliable parsing. It awards $1$ for correct formatting and $0$ otherwise.
    \item \textbf{Parameter-Agnostic Structural Reward ($R_{\text{structural}}$):} This is the core of our reward mechanism, evaluating the fundamental correctness of the generated law's structure. As detailed in Appendix~\ref{sec:reward_details}, it calculates the Jaccard similarity between the ``structural skeletons" of the generated ansatz and the canonical equation. This metric rewards topological correctness over superficial coefficient matching, aligning the optimization objective with the VIPER-R1's primary role.
    \item \textbf{Exact Match Accuracy Reward ($R_{\text{accuracy}}$):} This component provides the strictest evaluation, awarding a binary reward of $1$ only if the generated formula $S_i$ is symbolically identical to the ground truth $S_{\text{GT}}$. This encourages ultimate precision of the model's output.
\end{itemize}

\vspace{\baselineskip}
\noindent\textbf{Policy Update with Relative Advantage.}
The rewards $\{r_1, r_2, \dots, r_G\}$ for the group of sampled hypotheses are normalized to compute their relative advantages, preventing instability from high-variance rewards. The relative advantage $A_i$ for each ansatz $S_i$ is defined as:
\begin{equation}
    A_i = \frac{r_i - \text{mean}(r_1, \dots, r_G)}{\text{std}(r_1, \dots, r_G) + \epsilon}.
\end{equation}
The policy $\pi_{\theta}$ is then updated to increase the likelihood of generating hypotheses with positive advantages. This process is further regularized by a Kullback–Leibler divergence penalty between the updated policy and the original reference policy from the MSI stage, ensuring stable learning and preventing the model from deviating too far from its physically-grounded foundation.

\subsection{Agentic Refinement via Symbolic Residual Realignment (SR²)}

Upon completing its internal calibration, the VIPER-R1 has produced a high-confidence Symbolic Ansatz, denoted as $S_0$. This expression represents a robust, first-order approximation of the system's dynamics. In the final stage of our framework, the VIPER-R1 transitions into an \textit{agentic} role.
It recognizes that while its ansatz predicts a target variable $\hat{a}_{\text{VLM}}$, a discrepancy or ``residual field" may exist between this theoretical model and the precise empirical evidence.

To characterize and correct for this residual, the VIPER-R1 agentically invokes an external tool: a high-performance symbolic regression engine \citep{Cranmer2023Interpretable}. We term this sophisticated tool-use process \textbf{Symbolic Residual Realignment (SR²)}. This technique mirrors a physicist performing a perturbation analysis to account for higher-order effects, thereby realigning their theory with empirical reality.

\vspace{\baselineskip}
\noindent\textbf{The SR² Process.}
The core principle of SR² is to dramatically simplify the task for the symbolic regression tool. Instead of tasking it with searching the entire, near-infinite space of possible physical laws, we constrain its search to the much smaller, well-behaved space of the residual error. The process unfolds as follows:

\paragraph{Step 1: Residual Field Calculation:} The residual field, $r(t)$, is computed as the difference between the ground-truth target values from the empirical data, $a_{\text{GT}}(t)$, and the prediction from the VIPER-R1's Symbolic Ansatz:
    \begin{equation}
        r(t) = a_{\text{GT}}(t) - \hat{a}_{\text{VLM}}(x, v, t).
        \label{eq:residual}
    \end{equation}

\paragraph{Step 2: Symbolic Regression on the Residual:} The SR engine is then deployed with the explicit goal of finding a parsimonious and accurate symbolic expression, $S_{\text{residual}}$, that best models the residual field $r(t)$. This focused task allows the SR tool to operate with maximum efficiency.
    \begin{equation}
        a_{\text{residual}}(x, v, t) \leftarrow \text{SR}(x, v, t, r(t)).
    \end{equation}
    
\paragraph{Step 3: Theory Realignment:} The final, empirically-realigned Law of Motion, $S_{\text{final}}$, is constructed by composing the VIPER-R1's initial ansatz with the discovered residual expression. This yields a complete and highly accurate model of the system's dynamics.
    \begin{equation}
        a_{\text{final}}(x, v, t) = \hat{a}_{\text{VLM}}(x, v, t) + a_{\text{residual}}(x, v, t).
        \label{eq:final_formula}
    \end{equation}

The whole process of SR² is summarized in Algorithm \ref{alg:sr2_formatted}.

\section{Experiments}


\subsection{Experimental Setup}

\noindent\textbf{Dataset.} To avoid data contamination, where models might have seen common benchmarks during pre-training, we adopt the strategy of constructing a new synthetic dataset, as advocated in prior work such as LLM-SR~\citep{Shojaee2025LLMSR}. 
All experiments are conducted on our purpose-built \textbf{PhysSymbol} corpus, which contains 5,000 multimodal instances (a pair of kinematic plots and trajectory data files) each representing a unique, complex physical system. 
Further details on data generation and statistics are provided in Appendix~\ref{app:dataset}.

\vspace{\baselineskip}
\noindent\textbf{Models and Baselines.} Our primary models, \textbf{VIPER-R1-3B} and \textbf{VIPER-R1-7B}, are based on the Qwen-VL-2.5 3B and 7B architectures, respectively. 
We compare against a diverse set of state-of-the-art  MLMs, including \textit{GPT-5}~\citep{gpt5}, \textit{GPT-5 mini}~\citep{gpt5}, \textit{GPT-4o mini}~\citep{gpto4mini}, \textit{GPT-4o}~\citep{openai2024gpt4ocard}, \textit{Grok 3}~\citep{grok3}, \textit{GPT-o3}~\citep{gpto4mini}, \textit{Claude-4 Sonnet}~\citep{claude4sonnet}, \textit{Claude-3.7 Sonnet}~\citep{claude37},\textit{ Qwen-VL-Max}~\citep{Bai2023QwenVL}, \textit{Qwen-VL-2.5-72B-Instruct}~\citep{bai2025qwen2.5}, and\textit{ Gemini 2.5 Pro}~\citep{gemini25}.

\vspace{\baselineskip}
\noindent\textbf{Evaluation Metrics.} We evaluate the models across several dimensions to capture different aspects of performance:
\begin{itemize}[labelsep = .5em, leftmargin = 0pt, itemindent = 1em]
    \item \textbf{Structural Score ($S_{\text{struct}}$):} This is our primary metric for the VLM's hypothesis generation capability. It is the parameter-agnostic Jaccard similarity between the terms of the generated formula and the canonical equation. A score of 1.0 indicates a perfect structural match.
    \item \textbf{Accuracy Score ($S_{\text{acc}}$):} A stricter metric that measures the rate of exact symbolic matches between the generated formula and the canonical equation.
    \item \textbf{Post-SR² MSE:} The final Mean Squared Error of the complete, realigned formula after the \textbf{SR²} stage. This measures the end-to-end performance of the entire framework. A lower value is better.
\end{itemize}
Further experimental details, including model architectures, training procedures, and evaluation protocols, are provided in Appendix~\ref{app:train_evl}.

\subsection{Main Results and Analysis}
To validate the effectiveness of our approach, we benchmarked VIPER-R1 against a comprehensive suite of SOTA VLMs. The main results, as shown in Table~\ref{tab:main_results}, show that our framework outperforms other general-purpose VLMs.

\textbf{Superiority in Initial Hypothesis Generation.} The first two metrics, Structural Score ($S_{\text{struct}}$) and Accuracy Score ($S_{\text{acc}}$), evaluate the quality of the initial formula generated by the model before any symbolic refinement. Our specialized models demonstrate a commanding lead in this crucial first step. Our VIPER-R1-7B achieves an $S_{\text{struct}}$ of 0.812, representing a 56.7\% relative improvement over the best-performing baseline, Claude-4-Sonnet. Similarly, its $S_{\text{acc}}$ of 0.487 surpasses the top zero-shot model by over 45.4\%. These results shows that while capable of broad multimodal tasks, they lack the specialized reasoning abilities required to interpret the nuanced patterns of physical phenomena. Our two-stage training curriculum successfully imbues the model with this domain-specific, physicist-like intuition.

\textbf{Excellence in Final Law Discovery.} The ultimate goal is to find the most accurate physical law, a performance captured by the final Post-SR² MSE. A high-quality initial hypothesis from the VLM is critical, as it provides a much better starting point for the symbolic regression tool to find the true global optimum. Our results confirm this synergy. The superior initial guesses from VIPER-R1 lead to significantly more accurate final discoveries. VIPER-R1-7B model achieves a final MSE of only 0.032, an error rate nearly three times lower than the best baseline result of 0.091. It is noteworthy that even our smaller 3B model, with a final MSE of 0.081, outperforms all other SOTA VLMs.

\begin{table}[h!]
\centering
\caption{Main results comparing VIPER-R1 against SOTA VLMs on the PhysSymbol test set. Our method achieves the highest structural and accuracy scores, leading to the lowest final error.} 
\label{tab:main_results}
\resizebox{\columnwidth}{!}{%
\begin{tabular}{llccc}
\toprule
\textbf{Category} & \textbf{Method} & \textbf{Structural Score ($S_{\text{struct}}$) $\uparrow$} & \textbf{Accuracy Score ($S_{\text{acc}}$) $\uparrow$} & \textbf{Post-SR² MSE $\downarrow$} \\
\midrule
\multirow{3}{*}{VLMs} 
& GPT-5  & 0.494 & 0.363 & 0.192 \\
& GPT-5-mini  & 0.455 & 0.350 & 0.154 \\
& GPT-4o mini  & 0.463 & 0.235 & 0.109 \\
& GPT-4o  & 0.449 & 0.274 & 0.286 \\
& Grok 3  & 0.026 & 0.019 & 0.177 \\
& GPT-o3  & 0.502 & 0.335 & 0.234 \\
& Claude-4-Sonnet  & 0.518 & 0.257 & 0.091 \\
& Claude-3.7-Sonnet  & 0.485 & 0.294 & 0.136 \\
& Qwen-VL-Max  & 0.493 & 0.285 & 0.210 \\
& Qwen-VL-2.5-72B-Instruct  & 0.466 & 0.284 & 0.198 \\
& Gemini 2.5 Pro & 0.302 & 0.237 & 0.107 \\
\midrule
\multirow{2}{*}{\textbf{Ours}} 
& VIPER-R1-3B & 0.728 & \textbf{0.488} & 0.081 \\
& VIPER-R1-7B & \textbf{0.812} & 0.487 & \textbf{0.032} \\
\bottomrule
\end{tabular}%
}
\end{table}


\begin{figure}
    \centering
    \includegraphics[width=1\linewidth]{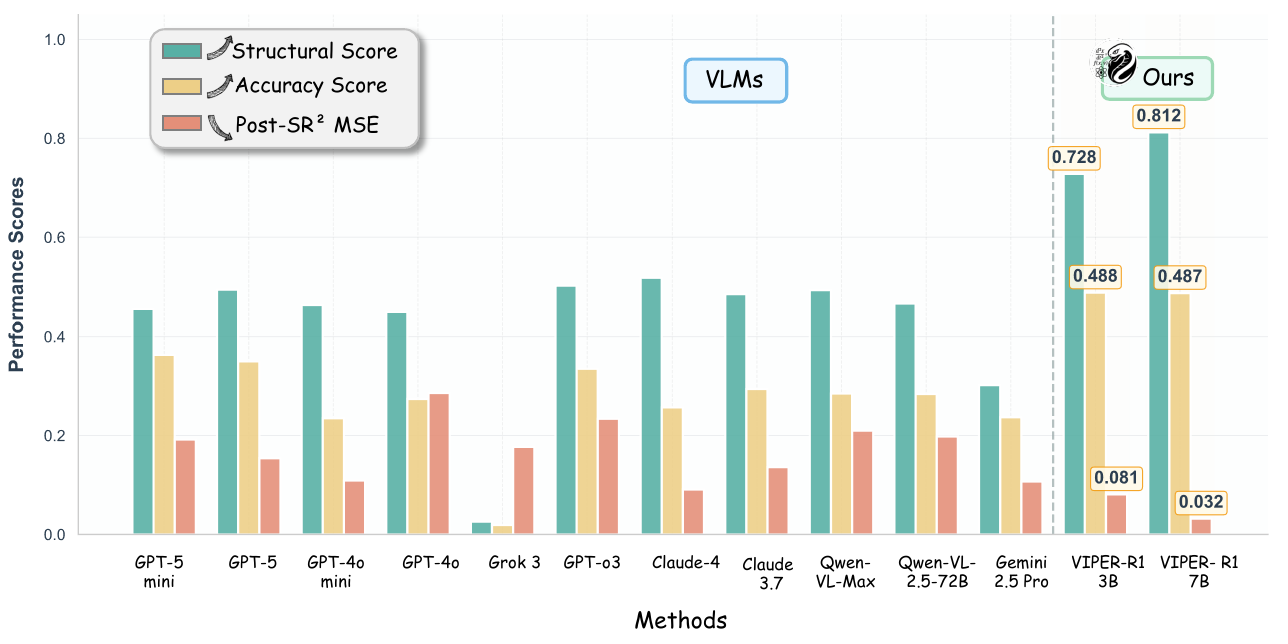}
    \caption{Quantitative comparison of model performance on the PhysSymbol test set. We report three metrics: structural score , accuracy score, and post-symbolic-regression MSE. VIPER-R1 (ours) outperforms all VLM baselines across all metrics, demonstrating significant improvements in both symbolic structure induction and predictive accuracy.}
    \label{fig:vlm_comparison}
\end{figure}

\section{Conclusion}
We introduced \textbf{\alias}, a multimodal framework that grounds symbolic reasoning in visual perception, emulating the scientific workflow. 
Through Motion Structure Induction and Reward-Guided Symbolic Calibration, the model generates robust symbolic hypotheses, while Symbolic Residual Realignment enables agentic refinement via external tools. 
Experiments on the PhysSymbol benchmark show that our multi-stage approach outperforms state-of-the-art VLMs, and traditional methods. 
Future directions include scaling to larger datasets, including chaotic systems and partial differential equations (PDEs), and extending from simulated plots to real experimental videos.


\bibliography{reference}
\bibliographystyle{iclr2025_conference}

\newpage
\appendix
\section{Appendix}

This supplementary material provides additional details on the proposed method and experimental results that could not be included in the main manuscript due to page limitations.
Specifically, this appendix is organized as follows.

\begin{itemize}
\item Sec.~\ref{app:train_evl} outlines the models, training processes, and more evaluation details,  providing more detailed experimental specifics.
\item Sec.~\ref{app:dataset} provides more details about PhysSymbol and discusses how we collected, filtered, and reconstructed a high-quality dataset.
\item Sec.~\ref{app:case} includes more visualization cases.
\end{itemize}

\section{Details of Training and Evaluation}
\label{app:train_evl}

\subsection{Training Settings}
We utilize Qwen2.5-VL-3B and Qwen2.5-VL-7B \citep{bai2025qwen2} as the backbone models for our experiments. Our implementation is built on the open-source frameworks Open-R1~\citep{openr1} and vLLM~\citep{vllm}, ensuring reproducibility and scalability. All experiments were conducted on a cluster of servers, each equipped with 8$\times$A800 GPUs. 
During MSI(SFT) stage, we train model for 5 epoch at each step. At RL refinement stage, the model is trained for 2 epoch.

\subsection{System Prompts} 
The behavior and reasoning process of VIPER-R1 are carefully guided by a series of structured system prompts tailored to each stage of our training and inference pipeline. These prompts define the model's role as a scientific assistant and establish the expected format for its reasoning and final output. This structured approach is crucial for decoupling complex tasks and progressively building the model's capability for scientific discovery. Below, we detail the specific prompts used in each phase.

\subsubsection{Prompt for MSI step 1}

\begin{figure}[ht]
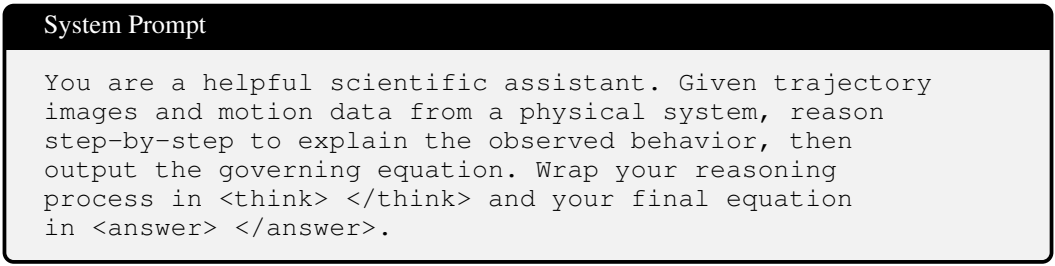

\centering
\begin{tcolorbox}[colback=gray!10, colframe=black, title=System Prompt, enhanced]
\begin{verbatim}
You are a helpful scientific assistant. Given trajectory 
images and motion data from a physical system, reason 
step-by-step to explain the observed behavior, then 
output the governing equation. Wrap your reasoning 
process in <think> </think> and your final equation 
in <answer> </answer>.
\end{verbatim}
\end{tcolorbox}
\caption{System prompt for the first SFT stage (MSI).}
\label{fig:prompt_sft1}
\end{figure}

In the initial MSI step, as shown in Figure \ref{fig:prompt_sft1},  the goal is to teach the model to perform end-to-end reasoning, connecting raw visual phenomena directly to a final governing equation. The prompt instructs the model to act as a scientific assistant, verbalize its step-by-step analysis, and provide a conclusive answer in a structured format.

\subsubsection{Prompt for MSI step 2}
In the second step of MSI, as shown in Figure \ref{fig:prompt_sft2},   we decouple the task: the model is provided with the pre-computed reasoning chain (C-CoT) and is tasked only with translating this analysis into a precise symbolic equation. This prompt focuses the model's training on the final, crucial step of symbolic formulation.

\begin{figure}[ht]
\centering
\begin{tcolorbox}[colback=gray!10, colframe=black, title=System Prompt, enhanced]
\begin{verbatim}
You are a helpful scientific assistant. Given the 
reasoning steps for a physical system and its 
trajectory images, output the corresponding governing 
equation. The reasoning is provided in <think> </think> 
tags, and your answer should be placed in 
<answer> </answer> tags.
\end{verbatim}
\end{tcolorbox}
\caption{System prompt for the second SFT stage (CoT-Aware).}
\label{fig:prompt_sft2}
\end{figure}

\subsubsection{Prompt for RGSC}
During the reinforcement learning phase, as shown in Figure \ref{fig:prompt_rgsc}, the prompt is refined to encourage more abstract and generalized symbolic reasoning. It explicitly asks the model to use symbolic placeholders for unknown parameters, which is essential for discovering general physical laws rather than fitting to specific numerical instances. This prompt guides the generation of hypotheses that are then evaluated by our reward function.

\begin{figure}[ht]
\centering
\begin{tcolorbox}[colback=gray!10, colframe=black, title=System Prompt, enhanced]
\begin{verbatim}
The user provides visual and trajectory data of a 
physical phenomenon. The Assistant's task is to act 
as a physicist. First, think step-by-step about the 
underlying physical principles in <think> tags. Then, 
derive and state the final governing equation in 
<answer> tags. The equation should use symbolic 
placeholders for unknown parameters (e.g., k, c, F) 
and standard variables for the system (x, v, t).
\end{verbatim}
\end{tcolorbox}
\caption{System prompt for the RGSC stage.}
\label{fig:prompt_rgsc}
\end{figure}

\subsection{Detailed Reward Function Formulation for RGSC}
\label{sec:reward_details}

The total reward signal $R(S_i)$ used during the Reward-Guided Symbolic Calibration (RGSC) stage is a weighted sum of three distinct components. Each component is designed to evaluate a specific aspect of the generated Symbolic Ansatz $S_i$, allowing for a balanced and effective policy optimization. The composite reward is defined as:
\begin{equation}
    R(S_i) = w_f R_{\text{format}}(S_i) + w_s R_{\text{structural}}(S_i, S_{\text{GT}}) + w_a R_{\text{accuracy}}(S_i, S_{\text{GT}}),
\end{equation}
where $S_{\text{GT}}$ is the Canonical Governing Equation, and $w_f, w_s, w_a$ are hyperparameters that weight the contribution of each reward component. Below, we detail the formulation of each component.

\paragraph{Format Reward ($R_{\text{format}}$).}
The primary purpose of this reward is to enforce a consistent and parsable output structure, which is crucial for both interpretability and automated evaluation. We use regular expressions to verify that the model's output strictly adheres to our predefined template, which requires a reasoning process enclosed within \texttt{<think>...</think>} tags followed by a final symbolic formula within \texttt{<answer>...</answer>} tags. This is a binary reward:
\begin{equation}
    R_{\text{format}}(S_i) = \begin{cases} 
        1 & \text{if format is correct} \\ 
        0 & \text{otherwise} 
    \end{cases}
\end{equation}

\paragraph{Parameter-Agnostic Structural Reward ($R_{\text{structural}}$).}
This is the most critical component for our task, designed to assess the fundamental topological correctness of the posited physical law. It rewards the model for identifying the correct basis functions and their relationships (e.g., $-k*x$, $-c*v$), irrespective of the specific values or symbols used for the coefficients (e.g., $k$, $c$). The calculation involves two steps:
\begin{enumerate}
    \item Both the generated ansatz $S_i$ and the ground truth $S_{\text{GT}}$ are parsed into symbolic expressions. We then decompose each expression into a set of constituent terms. For additive expressions, these are the terms separated by addition. For non-additive expressions, the term is the expression itself.
    \item Each term is then normalized into a ``structural skeleton" by replacing all numerical coefficients and symbolic parameters with a signed unit (i.e., $+1$ or $-1$), while preserving the core physical variables ($x, v, t$) and mathematical operators.
\end{enumerate}
The final reward is the Jaccard similarity between the set of skeletonized terms from the generated formula ($\mathcal{T}_{\text{gen}}$) and the ground truth ($\mathcal{T}_{\text{GT}}$). This provides a fine-grained score between 0 and 1.
\begin{equation}
    R_{\text{structural}}(S_i, S_{\text{GT}}) = \frac{|\mathcal{T}_{\text{gen}} \cap \mathcal{T}_{\text{GT}}|}{|\mathcal{T}_{\text{gen}} \cup \mathcal{T}_{\text{GT}}|}
\end{equation}

\paragraph{Exact Match Accuracy Reward ($R_{\text{accuracy}}$).}
This component provides the strictest evaluation, rewarding the model only if its generated symbolic formula is mathematically identical to the ground truth. This encourages ultimate precision, especially for formulas where parameters are also represented symbolically (e.g., using `k' instead of a number). We leverage the \texttt{sympy} library to perform a robust symbolic comparison. Both the generated answer and the ground truth are parsed into symbolic expressions, and we check if their simplified difference is zero. This is also a binary reward:
\begin{equation}
    R_{\text{accuracy}}(S_i, S_{\text{GT}}) = \begin{cases} 
        1 & \text{if } \texttt{sympy.simplify}(S_i - S_{\text{GT}}) = 0 \\ 
        0 & \text{otherwise} 
    \end{cases}
\end{equation}

By combining these three reward signals, we create a rich and nuanced optimization landscape. The model is primarily guided by the structural reward ($w_s$ is typically the largest weight) to learn the correct physics, while also being encouraged to produce well-formatted and, when possible, exactly correct symbolic expressions.

\subsection{Algorithm Pseudocode}

To provide a comprehensive and reproducible overview of our methodology, we present the detailed pseudocode for our framework's two primary components: the end-to-end training process and the inference-time refinement procedure. 

Algorithm~\ref{alg:training_framework} outlines the complete training framework for the VIPER-R1. This algorithm details the two primary phases through which the model is forged: first, the supervised Motion Structure Induction curriculum, which teaches the model to form hypotheses from visual data; and second, the subsequent reinforcement learning phase of Reward-Guided Symbolic Calibration, which purifies the model's symbolic generation policy.

Following this, Algorithm~\ref{alg:sr2_formatted} describes the inference procedure. This algorithm formalizes the Agentic Refinement via Symbolic Residual Realignment process, wherein the fully trained VIPER-R1 generates an initial hypothesis and then proactively invokes an external symbolic regression tool to produce a final, empirically-realigned physical law.

\begin{algorithm}[ht]
    \SetAlFnt{\small}
    \SetKwInOut{Inputs}{Inputs}
    \SetKwInOut{Outputs}{Outputs}
    \caption{VIPER-R1 Training Framework: MSI and RGSC}
    \label{alg:training_framework}
    \LinesNumbered
    \SetAlgoLined
    
    \Inputs{The PhysSymbol Corpus $\mathcal{D}$, \\
        Initial model parameters $\theta_0$ from a pre-trained VLM, \\
        Number of MSI steps $N_{\text{MSI-1}}, N_{\text{MSI-2}}$, \\
        Number of RGSC steps $N_{\text{RGSC}}$, \\
        GRPO group size $G$, \\
        Reward weights $w_f, w_s, w_a$
    }
    \Outputs{The final, calibrated VIPER-R1 policy $\pi_{\text{RGSC}}$}
    
    \vspace{0.2em}
    \hrule
    \vspace{0.2em}
    
    \tcp{Phase 1: Motion Structure Induction (MSI)}
    $\pi_{\theta} \leftarrow \text{InitializeModel}(\theta_0)$\;
    
    \vspace{0.2em}
    \tcp{Step 1.1: Joint Induction of C-CoT and Symbolic Structure}
    \For{$i = 1$ to $N_{\text{MSI-1}}$}{
        Sample a batch $(E, Y) \sim \mathcal{D}$, where $Y = (C, S)$\;
        Update $\theta$ by descending the gradient of $\mathcal{L}_{\text{MSI-1}}$ w.r.t. Eq.~\eqref{eq:msi1}\;
    }
    
    \vspace{0.2em}
    \tcp{Step 1.2: C-CoT-Guided Symbolic Formulation}
     \For{$i = 1$ to $N_{\text{MSI-2}}$}{
        Sample a batch $(E, C, S) \sim \mathcal{D}$\;
        Update $\theta$ by descending the gradient of $\mathcal{L}_{\text{MSI-2}}$ w.r.t. Eq.~\eqref{eq:msi2}\;
    }
    $\pi_{\text{MSI}} \leftarrow \pi_{\theta}$ 
    
    \vspace{0.2em}
    \hrule
    \vspace{0.2em}
    
    \tcp{Phase 2: Reward-Guided Symbolic Calibration (RGSC)}
    $\pi_{\theta} \leftarrow \pi_{\text{MSI}}$; $\pi_{\text{ref}} \leftarrow \pi_{\text{MSI}}$
    
    \For{$k = 1$ to $N_{\text{RGSC}}$}{
        Sample a batch of Empirical Evidence $E \sim \mathcal{D}$\;

        \For{$j = 1$ to $G$}{
             $S_j \sim \pi_{\theta}(S \mid E)$\;
        }

        \For{$j = 1$ to $G$}{
            $r_j \leftarrow w_f R_{\text{format}}(S_j) + w_s R_{\text{structural}}(S_j, S_{\text{GT}}) + w_a R_{\text{accuracy}}(S_j, S_{\text{GT}})$\;
        }

        $A \leftarrow \text{Normalize}(r_1, \dots, r_G)$\;
        
        Update $\theta$ using advantages $A$ and a KL penalty against $\pi_{\text{ref}}$\;
    }
    $\pi_{\text{RGSC}} \leftarrow \pi_{\theta}$\;
    
    \vspace{0.2em}
    \Return{$\pi_{\text{RGSC}}$};
\end{algorithm}

\begin{algorithm}[ht]
\SetAlFnt{\small}
\SetInd{0.5em}{0.5em} 
\SetKwInOut{Inputs}{Inputs}
\SetKwInOut{Outputs}{Outputs}
\caption{Agentic Refinement via Symbolic Residual Realignment (SR²)}
\label{alg:sr2_formatted}
\LinesNumbered
\SetAlgoLined

\Inputs{Trained VIPER-R1 policy $\pi_{\text{VIPER-R1}}$, \\
  Empirical Evidence $E=(\mathcal{I}, \mathcal{D})$, \\
  Symbolic Regression engine $\mathcal{SR}$
}
\Outputs{The final, realigned Law of Motion $S_{\text{final}}$}

\vspace{0.2em}
\hrule
\vspace{0.2em}

\tcp{Stage 1: VLM Hypothesis Generation}
$S_0 \leftarrow \text{GenerateAnsatz}(\pi_{\text{VIPER-R1}}, E)$; \\
$a_{\text{VLM}} \leftarrow \text{CompileFunction}(S_0)$ 

\vspace{0.2em}
\tcp{Stage 2: Residual Field Calculation}
$a_{\text{GT}}( x, v, t )\leftarrow \text{ExtractData}(E.\mathcal{D})$ 
$r \leftarrow a_{\text{GT}} - \hat{a}_{\text{VLM}}(x, v, t)$ 

\vspace{0.2em}
\tcp{Stage 3: Tool-Using for Residual Modeling}
$S_{\text{residual}} \leftarrow \mathcal{SR}(\text{inputs}=(x,v,t), \text{target}=r)$ 

\vspace{0.2em}
\tcp{Stage 4: Theory Realignment}
$S_{\text{final}} \leftarrow S_0 + S_{\text{residual}}$ 

\vspace{0.2em}
\Return{$S_{\text{final}}$};
\end{algorithm}


\subsection{Evaluation Metrics}
\label{app:metrics}

To provide a holistic assessment of our framework's performance, we employ a suite of distinct metrics, each designed to capture a different facet of success, from high-level structural correctness to final empirical accuracy.

\paragraph{Structural Score ($S_{\text{struct}}$)}
This is our primary metric for evaluating the core capability of the VIPER-R1: its ability to generate a topologically correct symbolic hypothesis. This score measures the structural similarity between the generated formula and the canonical equation, intentionally ignoring numerical coefficients to focus purely on the underlying physical structure.


\paragraph{Accuracy Score ($S_{\text{acc}}$)}
To measure the exactness of the generated formulas, we use a strict symbolic accuracy score. This metric evaluates whether the generated formula is mathematically identical to the canonical equation. It serves as a challenging measure of the model's ultimate precision.


\paragraph{Post-SR² Mean Squared Error (MSE)}
This metric evaluates the end-to-end performance of the entire VIPER-R1 framework by measuring how well the final, refined formula fits the observed data. It quantifies the empirical accuracy after the \textbf{SR²} stage has been completed.

\noindent\textit{Calculation:} Let $S_{\text{final}}$ be the final symbolic law produced by our framework. This expression is converted into a callable function $a_{\text{final}}(x,v,t)$. The MSE is then computed over the $N$ data points in the test set's trajectory data:
\begin{equation}
    \text{MSE} = \frac{1}{N} \sum_{i=1}^{N} \left( a_{\text{GT}}(t_i) - a_{\text{final}}(x_i, v_i, t_i) \right)^2,
\end{equation}
where $a_{\text{GT}}(t_i)$ is the ground-truth acceleration at time $t_i$. A lower MSE indicates a better fit to the observed physical reality and thus a more successful discovery.

\subsection{Ablation Studies}
To dissect and quantify the contribution of each core component of our framework, we conducted a series of ablation studies on both the 3B and 7B model sizes. We systematically evaluate the performance of: (i) the base Qwen-VL-2.5 model, (ii) the model after only the SFT-based Motion Structure Induction stage, and (iii) our full model, which includes the subsequent RL-based Reward-Guided Symbolic Calibration stage. The results are presented in Table~\ref{tab:ablation}.

The results in Table~\ref{tab:ablation} reveal several findings. First, applying MSI alone yields a substantial performance boost over the base model—improving structural scores by over 40 points, which confirms that our two-stage SFT process effectively grounds symbolic reasoning in visual perception and physical intuition. Second, the addition of RGSC further elevates performance across both metrics. For instance, the 7B model’s structural score improves from 0.554 (MSI-only) to 0.812 after applying RGSC, and its accuracy score increases from 0.399 to 0.487. Similar trends are observed in the 3B model. These improvements highlight the importance of RL-based symbolic calibration: by optimizing outputs through reward-guided refinement, the model learns to produce more structurally sound and numerically accurate symbolic expressions.

\begin{table}[h!]
\centering
\caption{Ablation study on the contribution of MSI and RGSC stages for both 3B and 7B models. Each stage provides a significant performance boost.}
\label{tab:ablation}
\small
\begin{tabular}{llcc}
\toprule
\textbf{Model Size} & \textbf{Model Version} & \textbf{Structural Score ($S_{\text{struct}}$) $\uparrow$} & \textbf{Accuracy Score ($S_{\text{acc}}$) $\uparrow$} \\
\midrule
\multirow{3}{*}{7B} 
& Qwen-VL-2.5 (Base) & 0.096 & 0.179 \\
& + MSI (SFT only) & 0.554 & 0.399 \\
& \textbf{+ MSI + RGSC (Ours)} & \textbf{0.812} & \textbf{0.487} \\
\midrule
\multirow{3}{*}{3B} 
& Qwen-VL-2.5 (Base) & 0.043 & 0.100 \\
& + MSI (SFT only) & 0.474 & 0.350 \\
& \textbf{+ MSI + RGSC (Ours)} & \textbf{0.728} & \textbf{0.488} \\
\bottomrule
\end{tabular}
\end{table}

\begin{figure}[htbp]
    \centering
    \begin{subfigure}[t]{0.48\linewidth}
        \centering
        \includegraphics[width=\linewidth]{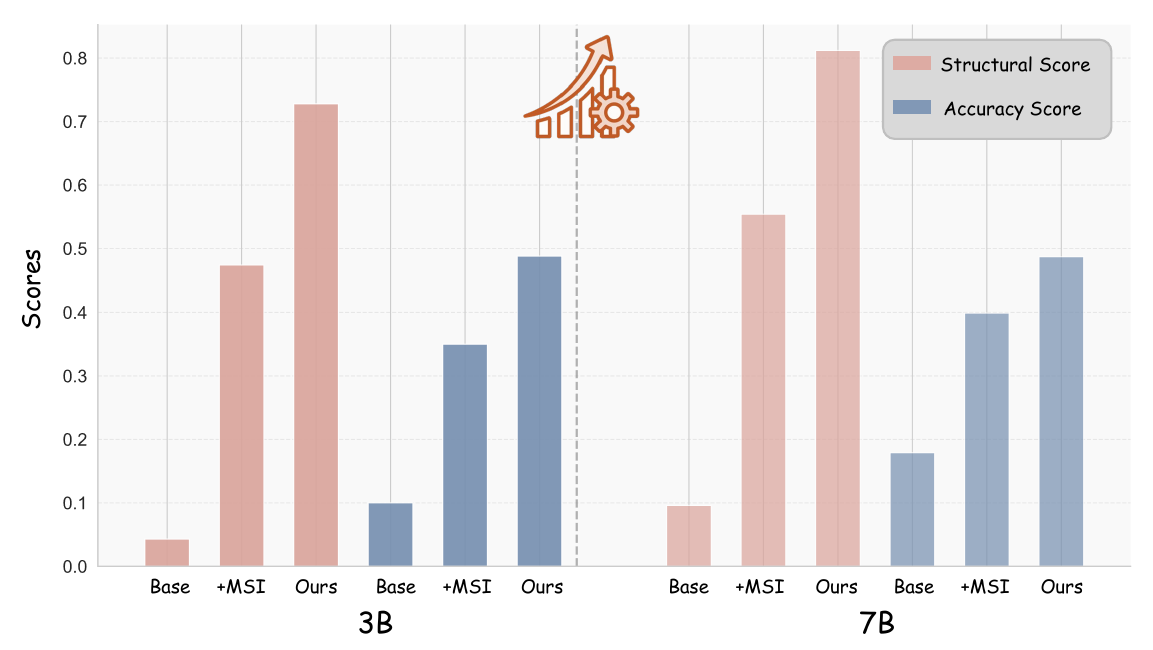}
        \label{fig:ablation}
    \end{subfigure}
    \hfill
    \begin{subfigure}[t]{0.48\linewidth}
        \centering
        \includegraphics[width=\linewidth]{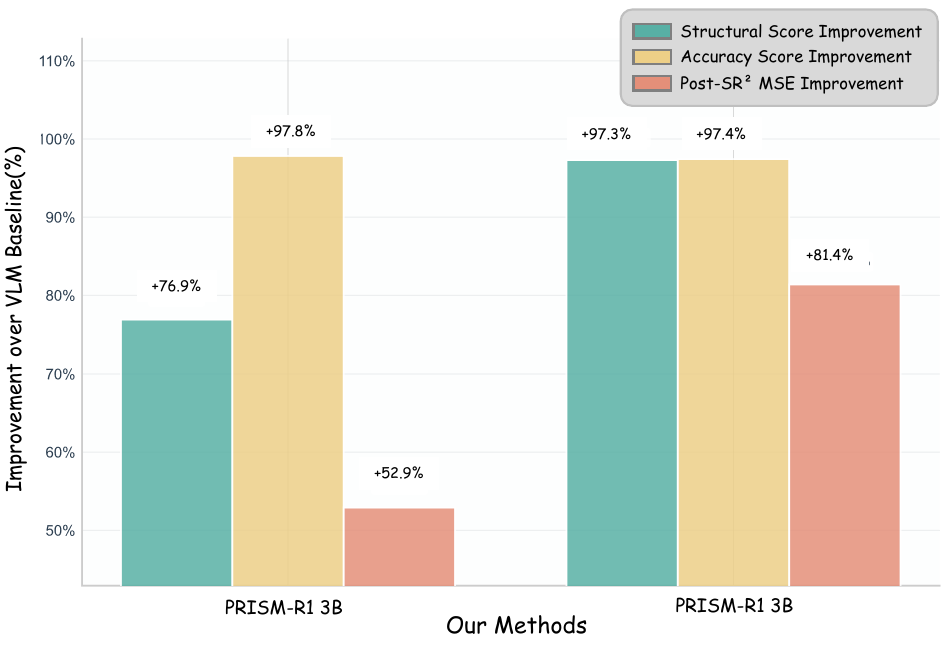}
        \label{fig:bar}
    \end{subfigure}
    \caption{(a) Ablation results show that integrating symbolic regression (+SR²) and our full VIPER-R1 pipeline progressively improves structural and accuracy scores. (b) VIPER-R1 achieves significant relative improvements over zero-shot VLM baselines across all evaluation metrics.}
    \label{fig:performance_combined}
\end{figure}

\section{PhysSymbol: A Comprehensive Multimodal Dataset for Physics Formula Discovery}
\label{app:dataset}

\subsection{Dataset Overview and Motivation}

To train and evaluate our proposed VIPER-R1 framework, we constructed \textbf{PhysSymbol}, a large-scale synthetic multimodal dataset that systematically emulates the analytical workflow of physicists studying complex dynamical systems. The dataset addresses a critical gap in existing benchmarks by providing paired visual-symbolic representations that capture the phenomenological patterns essential for scientific reasoning.

PhysSymbol comprises 5,000 instances, each containing a complete multimodal representation of a physical system: (1) dual trajectory visualizations (phase-space and temporal plots), (2) high-resolution numerical trajectory data, (3) ground-truth governing equations, and (4) expert-level causal reasoning annotations. This comprehensive design enables our framework to learn the crucial mapping from visual observations to symbolic mathematical expressions that characterizes human scientific discovery.

\subsection{Physics Term Library and Formula Generation}

The foundation of PhysSymbol lies in a carefully designed physics term library that encompasses the fundamental mechanisms commonly encountered in classical mechanics and nonlinear dynamics. Our term library includes 11 distinct categories of physical phenomena:

\textbf{Linear and Nonlinear Restoring Forces:}
\begin{itemize}[labelsep = .5em, leftmargin = 0pt, itemindent = 1em]
    \item Linear elasticity: $-kx$ with $k \in [0.1, 10]$
    \item Cubic nonlinearity: $-\beta x^3$ with $\beta \in [0.01, 5]$  
    \item Quintic nonlinearity: $-\delta x^5$ with $\delta \in [0.001, 1]$
\end{itemize}

\textbf{Velocity-Dependent Damping:}
\begin{itemize}[labelsep = .5em, leftmargin = 0pt, itemindent = 1em]
    \item Linear damping: $-cv$ with $c \in [0.01, 2]$
    \item Cubic velocity damping: $-\alpha v^3$ with $\alpha \in [0.01, 5]$
    \item Quintic velocity damping: $-\eta v^5$ with $\eta \in [0.001, 1]$
\end{itemize}

\textbf{External and Coupling Forces:}
\begin{itemize}[labelsep = .5em, leftmargin = 0pt, itemindent = 1em]
    \item Temporal periodic forcing: $F\sin(\omega t)$ with $F \in [0.1, 5]$, $\omega \in [0.5, 5]$
    \item Spatial periodic forcing: $F\sin(\omega x)$ with parameters in similar ranges
    \item Position-velocity coupling: $-\gamma xv$ with $\gamma \in [0.01, 5]$
\end{itemize}

\textbf{Specialized Nonlinear Terms:}
\begin{itemize}[labelsep = .5em, leftmargin = 0pt, itemindent = 1em]
    \item Trigonometric nonlinearity: $-x\cos(x)$, $-x\sin(x)$ (parameter-free)
    \item Stochastic perturbations: $\sigma \mathcal{N}(0,1)$ with $\sigma \in [0.01, 0.5]$
\end{itemize}

The formula generation process employs a structured combinatorial approach. Each governing equation is constructed by sampling 2-5 terms from the library, with a mandatory linear restoring force to ensure physical stability. Parameters are sampled uniformly from their respective ranges, and the resulting symbolic expression is converted into an executable function for numerical integration.

\subsection{High-Fidelity Trajectory Simulation}

For each generated governing equation of the form $\ddot{x} = f(x, \dot{x}, t)$, we perform high-resolution numerical integration using the adaptive Runge-Kutta method implemented in \texttt{scipy.integrate.solve\_ivp}. The simulation protocol follows these specifications:

\textbf{Temporal Parameters:}
\begin{itemize}[labelsep = .5em, leftmargin = 0pt, itemindent = 1em]
    \item Integration duration: $T = 20$ time units
    \item Sampling resolution: $N = 1000$ uniformly spaced points
    \item Time step: $\Delta t = 0.02$ (adaptive refinement as needed)
\end{itemize}

\textbf{Initial Conditions:}
Initial position $x_0$ and velocity $v_0$ are independently sampled from uniform distributions over $[-1, 1]$ to ensure diversity in trajectory patterns while maintaining numerical stability.

\textbf{Data Output:} Each simulation yields a trajectory dataset $\{(t_i, x_i, v_i, a_i)\}_{i=1}^N$ containing temporal evolution of position, velocity, and acceleration. This data is exported as CSV files with full numerical precision for downstream analysis.

\subsection{Dual Visualization Strategy}

A key innovation of PhysSymbol is its systematic generation of complementary visualizations that capture different aspects of system dynamics, mirroring the analytical tools used by practicing physicists:

\textbf{Phase-Space Portraits ($v$ vs $x$):} These plots encode the kinematic structure and stability properties of the dynamical system. Phase portraits reveal crucial qualitative features such as:
\begin{itemize}[labelsep = .5em, leftmargin = 0pt, itemindent = 1em]
    \item Closed orbits indicating conservative dynamics
    \item Spiral trajectories suggesting damped oscillations  
    \item Multiple attractors or limit cycles in nonlinear systems
    \item Geometric signatures of different restoring force types
\end{itemize}

\textbf{Temporal Trajectories ($x$ vs $t$):} These plots emphasize the time-domain behavior and temporal patterns:
\begin{itemize}[labelsep = .5em, leftmargin = 0pt, itemindent = 1em]
    \item Oscillation frequencies and amplitude modulation
    \item Exponential growth or decay envelopes
    \item Periodic forcing signatures and resonance effects
    \item Transient dynamics and approach to steady states
\end{itemize}

Both visualizations are rendered as high-resolution PNG images (300 DPI) with consistent styling, axis labeling, and grid structures to ensure visual uniformity across the dataset.

\subsection{Expert-Level Reasoning Annotation}

A key challenge in automated scientific discovery lies in bridging the gap between raw visual observation and symbolic reasoning. 
To address this, the \textbf{PhysSymbol} corpus incorporates detailed Causal Chain-of-Thought (C-CoT) annotations, designed to emulate the step-by-step reasoning of a human physicist. 
These annotations are generated through a carefully engineered prompting strategy with GPT-4o, implemented in \texttt{physics\_cot\_generator.py}, ensuring both consistency and expert-level interpretability.

For each physical system, the annotation protocol provides GPT-4o with both trajectory visualizations and the ground-truth governing equation, along with a structured prompt that explicitly requests analysis of visual-symbolic correspondences. 
This setup enables the model to reason not only about the equation itself, but also about how its individual terms manifest within the observed dynamical patterns. 

The resulting C-CoT traces follow a systematic analytical framework. 
They begin with \textbf{visual pattern recognition}, identifying salient geometric and temporal features in the trajectory and phase-space plots. 
This is followed by \textbf{physical interpretation}, where observed patterns are linked to underlying mechanisms such as oscillations, damping, or nonlinear effects. 
The annotations then provide \textbf{term-by-term analysis}, explaining how each component of the governing equation contributes to specific visual signatures. 
Building on this, they articulate a \textbf{hypothesis formation} step, reasoning about which terms could be posited from visual evidence alone. 
Finally, the traces include a layer of \textbf{validation logic}, explicitly connecting the proposed symbolic structure back to empirical observations for plausibility checking.

By design, these annotations capture expert-level reasoning that mirrors how human physicists transition from qualitative observation to quantitative formulation. 
They not only provide supervision for training multimodal models, but also serve as a valuable benchmark for evaluating whether models can perform reasoning that is genuinely interpretable and scientifically grounded.

\subsection{Dataset Assembly and Multi-Format Generation}

The final dataset assembly process, implemented in \texttt{build\_dataset\_json.py}, integrates all components into a unified multimodal format suitable for different training stages:

\textbf{Data Instance Structure:} Each complete instance follows the tuple format:
\[
(\text{Images: } I_{\text{phase}}, I_{\text{trajectory}}, \text{ Trajectory Data: } M, \text{ Ground-Truth Formula: } E, \text{ C-CoT Reasoning: } C)
\]

\textbf{Multi-Stage Training Variants:} To support our three-stage training pipeline, the assembly process generates three dataset variants:

\begin{enumerate}[labelsep = .5em, leftmargin = 0pt, itemindent = 1em]
    \item \textbf{Stage 1 (MSI-Joint):} Full format requiring both reasoning generation and formula prediction
    \item \textbf{Stage 2 (MSI-Guided):} C-CoT provided as input, only formula prediction required  
    \item \textbf{Stage 3 (RGSC):} Streamlined format for reinforcement learning with structural rewards
\end{enumerate}

\textbf{Data Preprocessing:} To ensure computational efficiency, trajectory data is intelligently subsampled to 100 uniformly spaced points while preserving essential dynamical characteristics. This balances information retention with processing speed during training.

\subsection{Dataset Statistics and Quality Assurance}

The \textbf{PhysSymbol} corpus consists of 5,000 complete multimodal instances, each pairing symbolic formulas with corresponding trajectory data and kinematic plots. 
The formulas exhibit controlled complexity, ranging from two to five terms with an average of 3.2 terms per equation. 
To ensure comprehensive coverage, parameters for all term types are fully sampled within their defined ranges. 
On the visual side, the dataset encompasses a diverse set of phase portrait topologies, capturing a wide variety of dynamical behaviors.

To guarantee reliability and usability, we apply a series of quality control measures. 
Each generated equation undergoes numerical stability verification to avoid degenerate or divergent solutions. 
Representative samples are visually inspected to confirm the clarity and readability of the plots. 
The accompanying Causal Chain-of-Thought (C-CoT) rationales are validated through automated keyword analysis to ensure coherence with the underlying formulas. 
Finally, file integrity checks are conducted across all multimodal components, ensuring that the dataset is complete, consistent, and ready for large-scale experimentation.

\section{Case Analysis}
\label{app:case}
In this section, we present detailed qualitative results for several challenging physical systems to provide a more intuitive understanding of the VIPER-R1 framework's capabilities. We first provide an in-depth analysis of a complex non-linear system (Case 1) and then present the visual results for three additional, distinct cases.

\subsection{In-Depth Analysis: Non-linear Damping with Stochastic Noise}

To showcase our method's full capabilities, we first examine a complex system governed by a linear restoring force, a non-linear damping term, and stochastic noise. The canonical governing equation is of the form $a(t) = -kx - cv^3 + \eta(t)$, where $\eta(t)$ represents a random noise component. This type of system is particularly challenging as it requires identifying and integrating components with fundamentally different mathematical and physical characteristics.

\begin{figure}[h!]
    \centering
    \includegraphics[width=\linewidth]{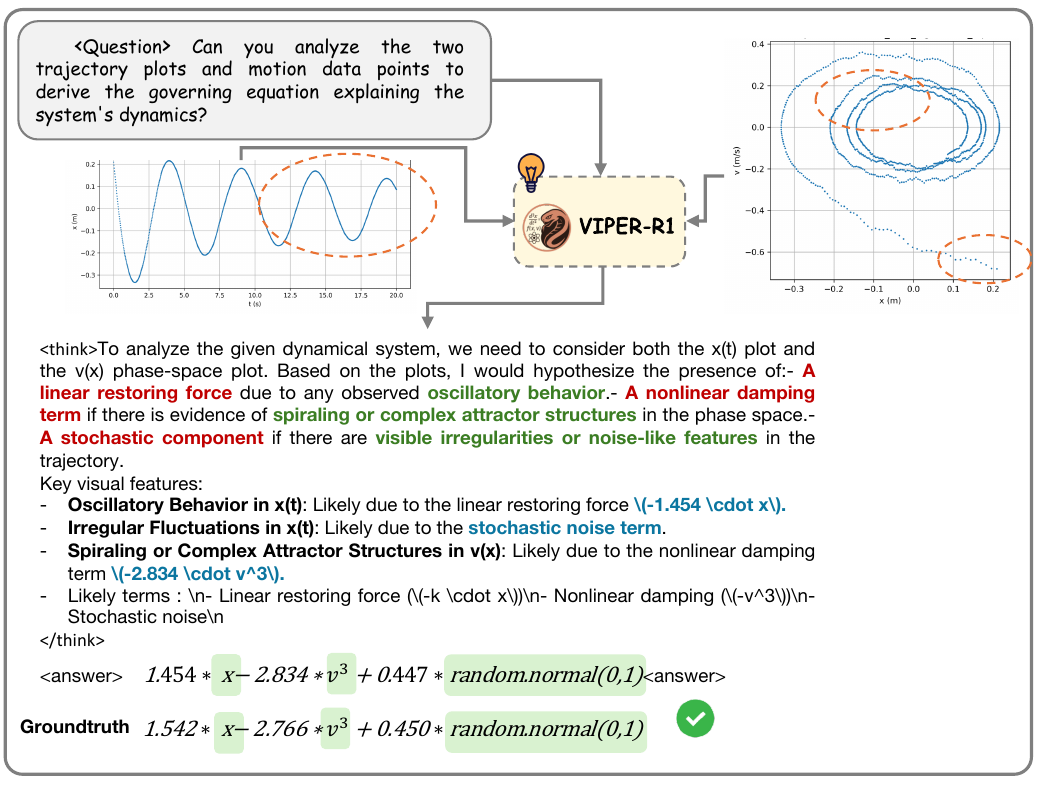}
    \caption{Case 1 Reasoning Process: The VIPER-R1 infers the governing equation of a non-linear dynamical system. Given the x(t) and v(x) plots, the model performs structured visual reasoning to identify key dynamics, including the linear restoring force, non-linear damping, and stochastic noise, before outputting an interpretable symbolic equation.}
    \label{fig:case_1_reasoning}
\end{figure}

As illustrated in Figure~\ref{fig:case_1_reasoning}, our VIPER-R1 leverages its C-CoT process to perform a sophisticated, physicist-like workflow. It correctly identifies distinct visual cues from the provided plots and maps each one to its underlying physical term.

The model's internal reasoning, extracted from its output, is as follows:
\begin{quote}
\texttt{<think>}
To analyze the given dynamical system, we consider both the $x(t)$ plot and the $v(x)$ phase-space plot.
\textbf{Key visual features:}
\begin{itemize}
    \item \textbf{Oscillatory Behavior in x(t):} Suggests a linear restoring force ($-k \cdot x$).
    \item \textbf{Distorted, Spiraling Attractor in v(x):} Indicates a non-linear damping term, likely dependent on a higher power of velocity ($-c \cdot v^3$).
    \item \textbf{Irregular, High-Frequency Fluctuations in x(t):} Points to a stochastic noise term.
\end{itemize}
\textbf{Conclusion:} The system likely combines a linear restoring force, non-linear damping ($-v^3$), and stochastic noise.
\texttt{</think>}
\end{quote}

\begin{figure}[h!]
    \centering
    \includegraphics[width=\linewidth]{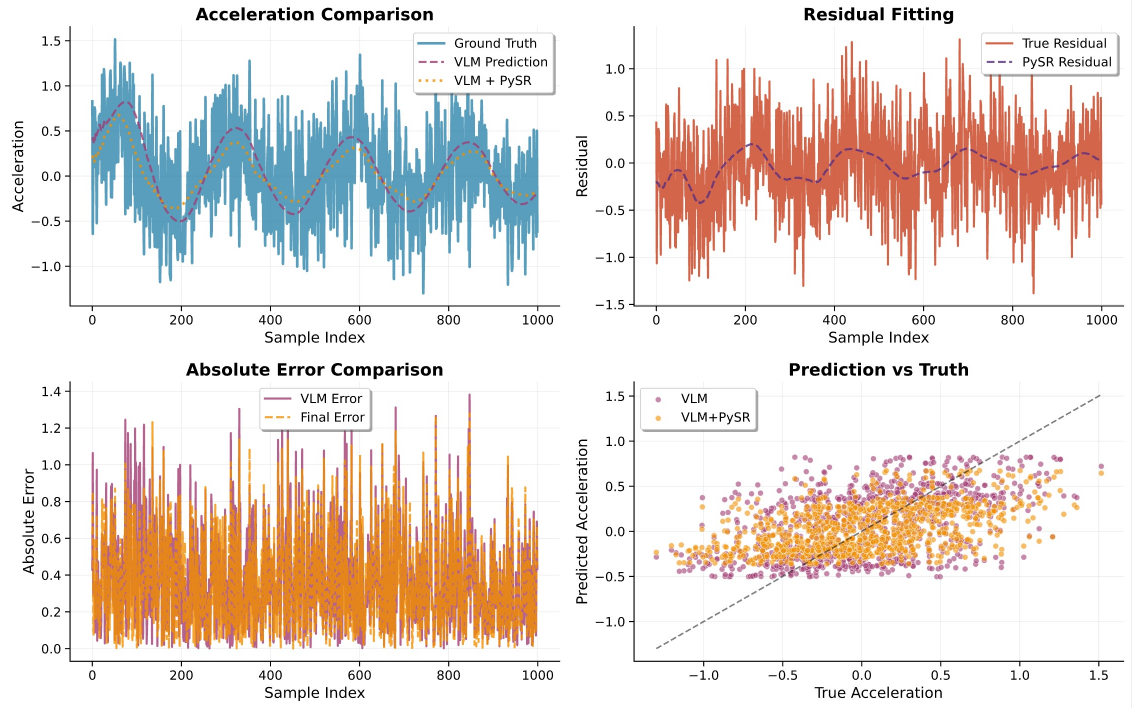}
    \caption{Case 1 Acceleration Signal Evaluation: Comparison of the predicted acceleration signals before (VLM-only) and after (VLM + SR²) symbolic refinement. The refined result demonstrates significantly improved alignment with the ground truth, as shown by the reduced residuals and errors.}
    \label{fig:case_1_acceleration}
\end{figure}

\begin{figure}[h]
    \centering
    \includegraphics[width=\linewidth]{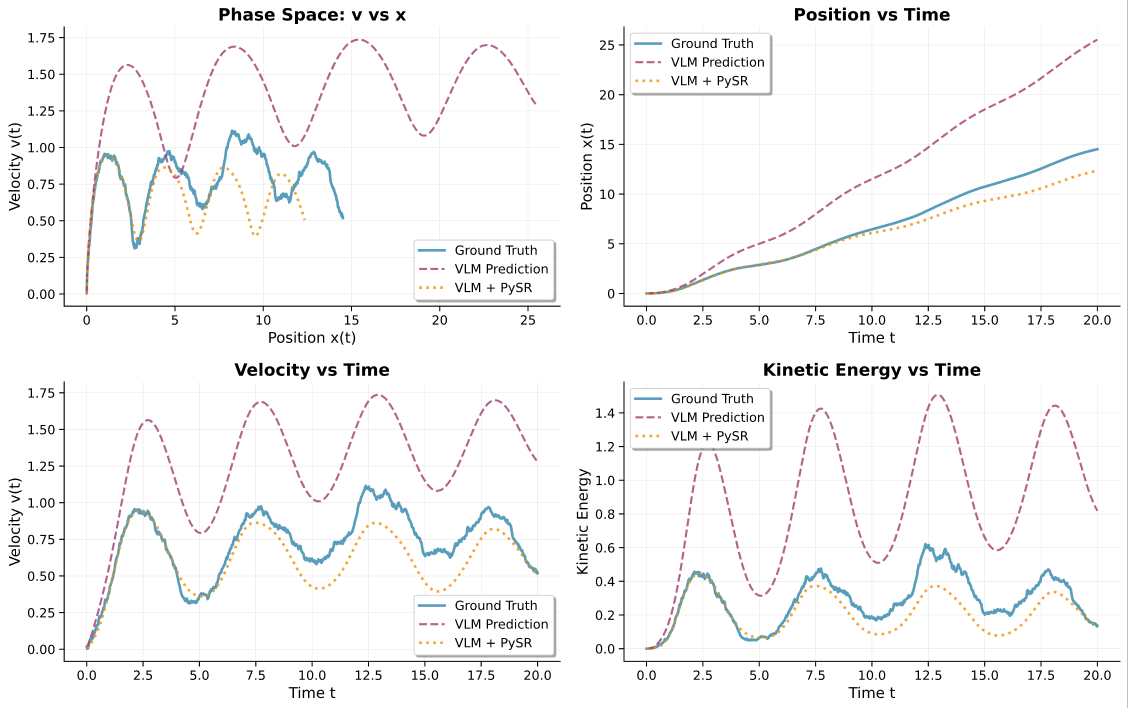}
    \caption{Case 1 System-Level Trajectory Comparison: The phase-space, position, velocity, and energy trajectories generated by the final (VLM + SR²) model show much closer adherence to the ground truth dynamics compared to the raw VLM predictions, indicating physically consistent long-term behavior.}
    \label{fig:case_1_trajectory}
\end{figure}

This detailed analysis leads the model to propose a hypothesis that is not only structurally correct but also quantitatively close to the true solution, providing an excellent starting point for the subsequent SR² stage. The quantitative success of this process is detailed in Figure~\ref{fig:case_1_acceleration} and Figure~\ref{fig:case_1_trajectory}, which show the improvements at both the signal and system levels.

\clearpage
\subsection{Additional Qualitative Results}
To further demonstrate the robustness of our framework, we present the results for three additional cases, each representing a different class of physical system.

\begin{figure}[h]
    \centering
    \includegraphics[width=\linewidth]{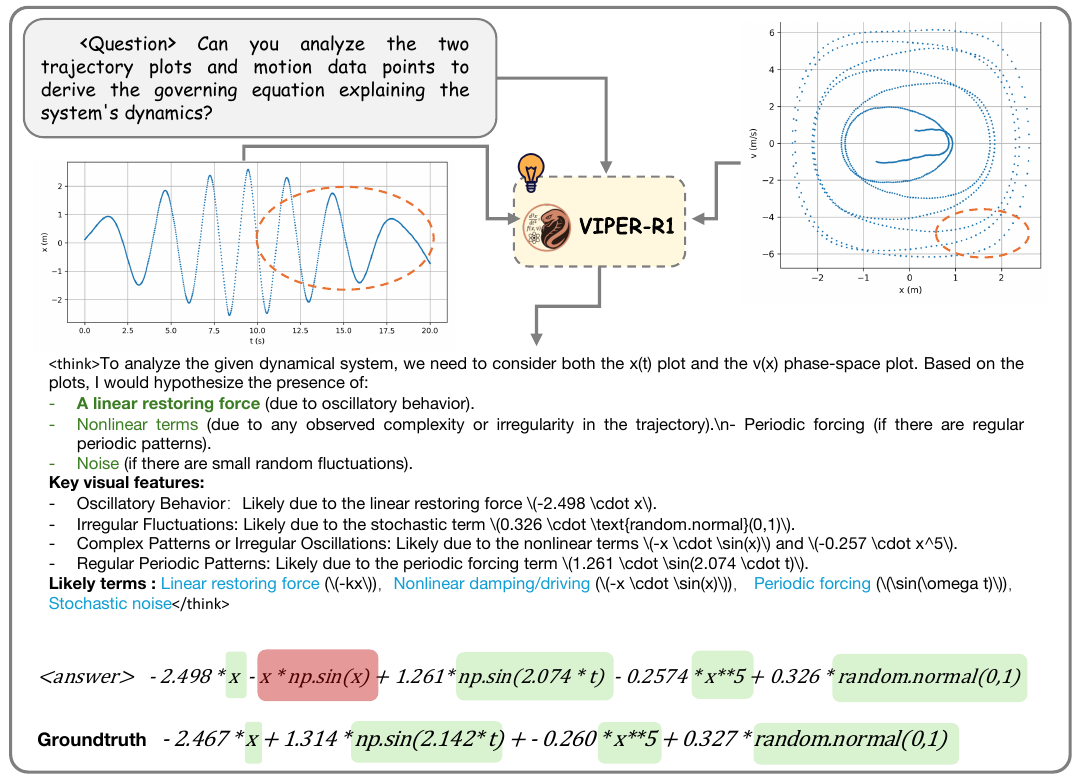}
    \caption{Case 2 Reasoning: A system with linear restoring forces.}
    \label{fig:case_2_reasoning}
\end{figure}

\begin{figure}[h]
    \centering
    \includegraphics[width=1\columnwidth]{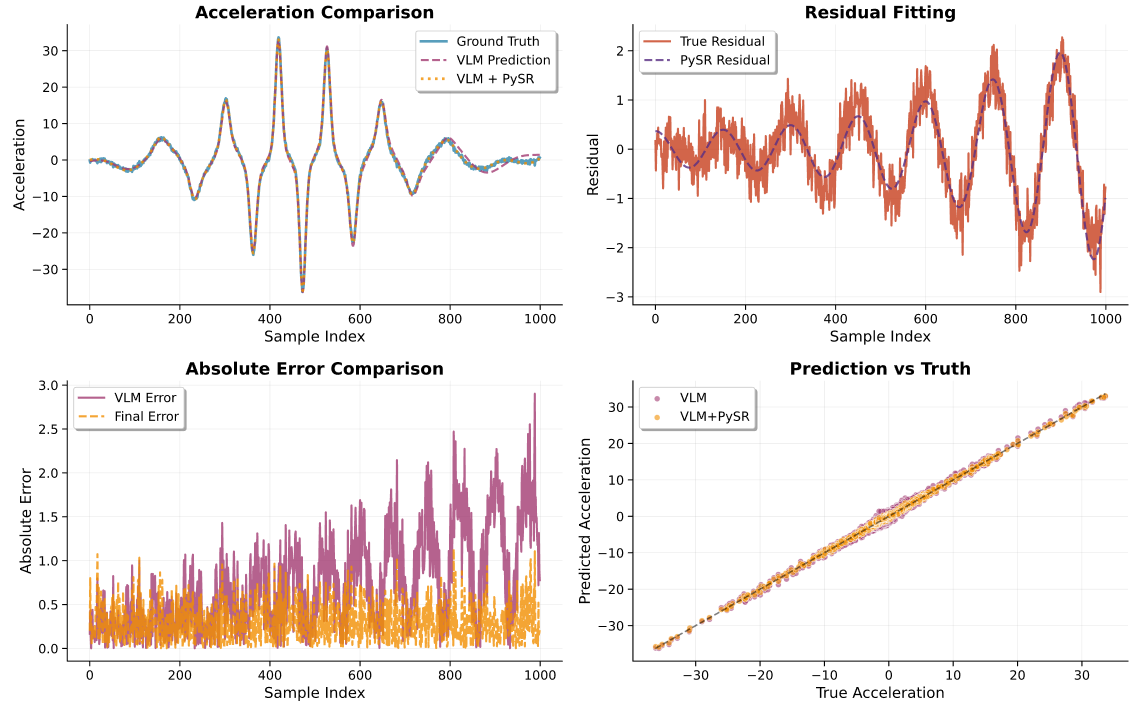}
    \caption{Case 2 Acceleration Signal Evaluation.}
    \label{fig:case_2_acceleration}
\end{figure}

\begin{figure}[h]
    \centering
    \includegraphics[width=1\columnwidth]{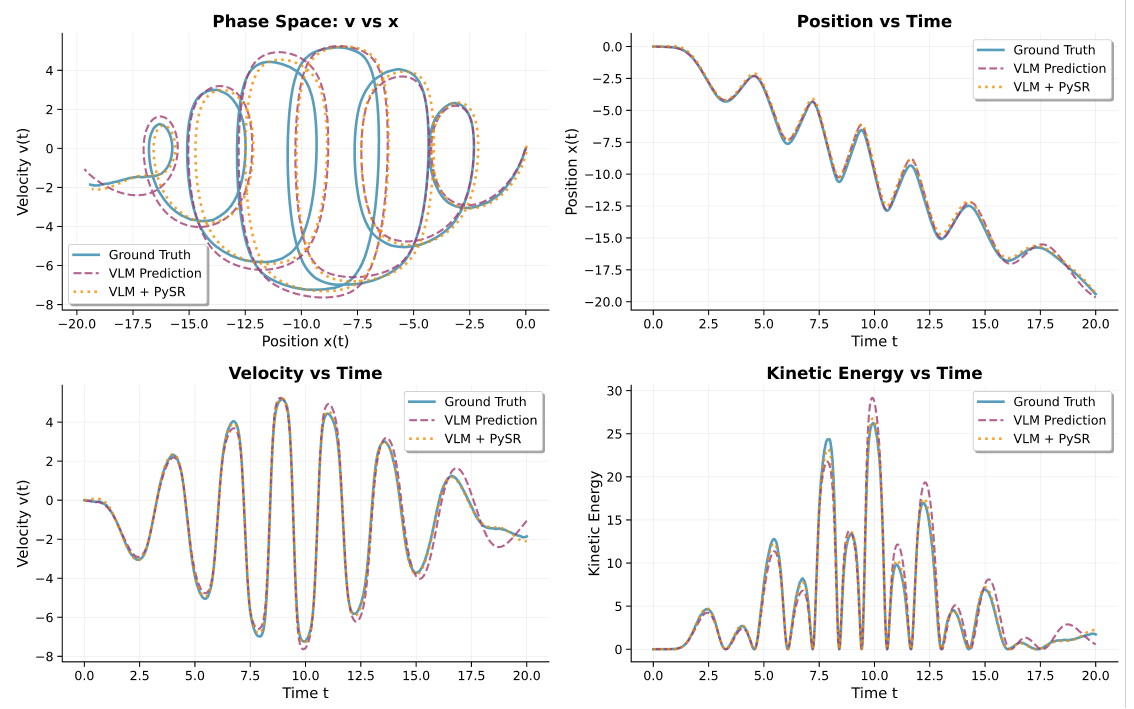}
    \caption{Case 2 System-Level Trajectory Comparison.}
    \label{fig:case_2_trajectory}
\end{figure}

\clearpage

\begin{figure}[h]
    \centering
    \includegraphics[width=\linewidth]{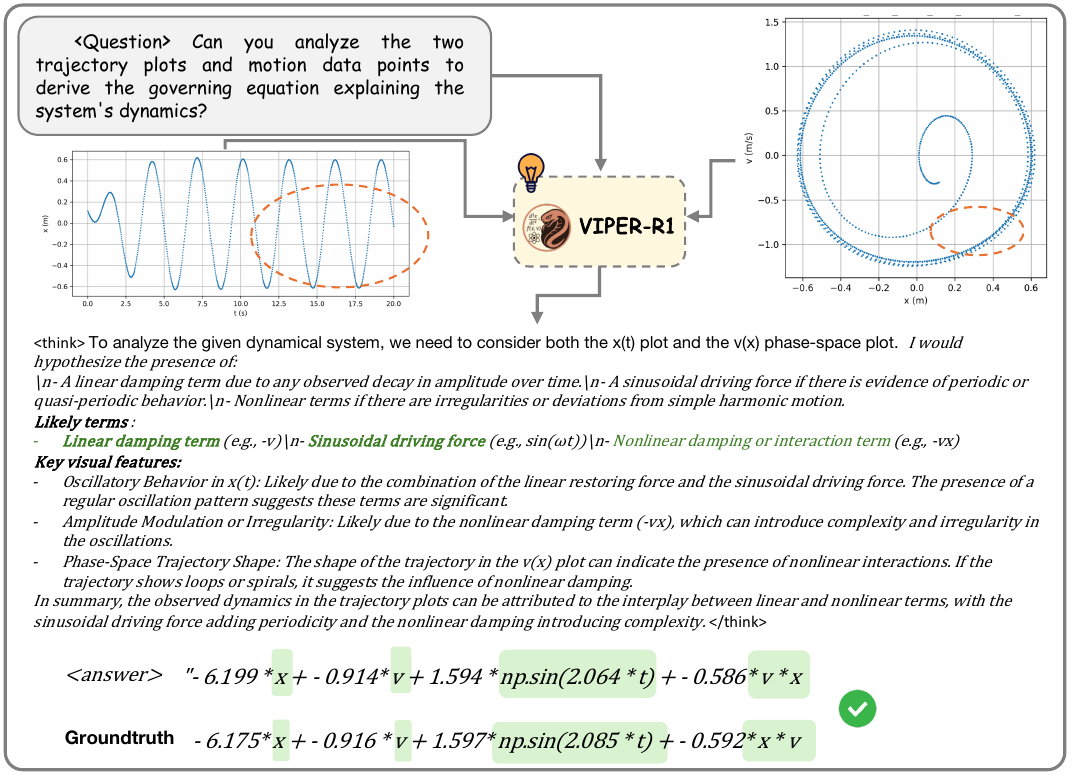}
    \caption{Case 3 Reasoning: A system with sinusoidal driving forces.}
    \label{fig:case_3_reasoning}
\end{figure}

\begin{figure}[h]
    \centering
    \includegraphics[width=1\columnwidth]{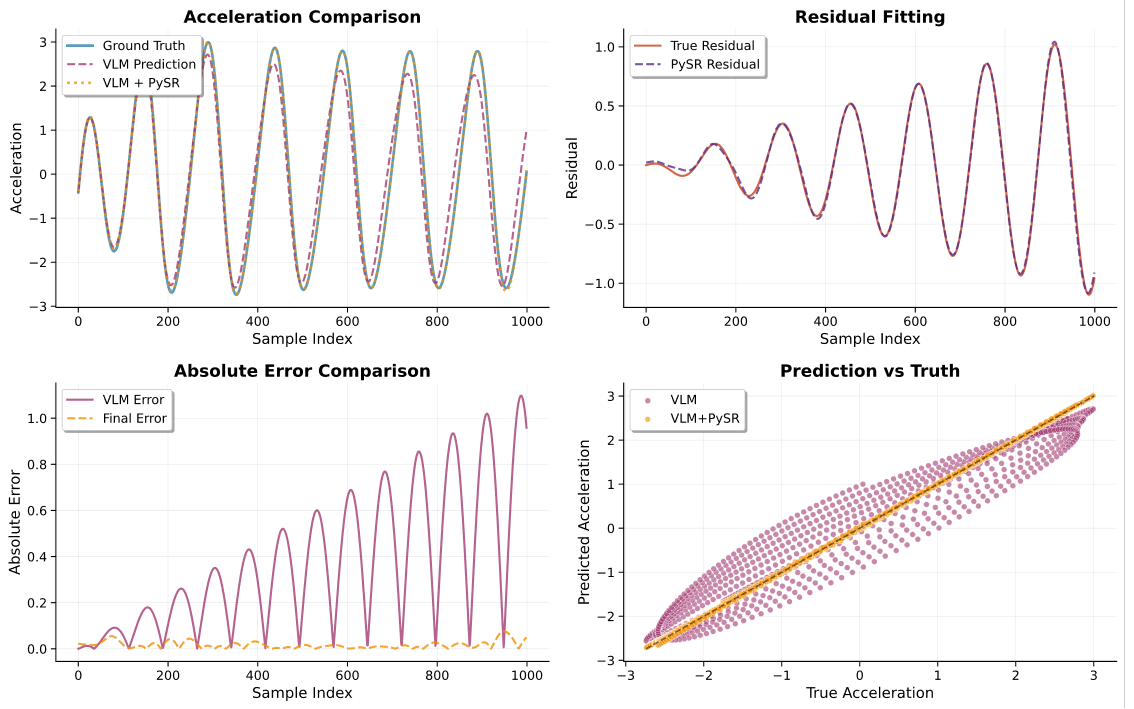}
    \caption{Case 3 Acceleration Signal Evaluation.}
    \label{fig:case_3_acceleration}
\end{figure}

\begin{figure}[h]
    \centering
    \includegraphics[width=1\columnwidth]{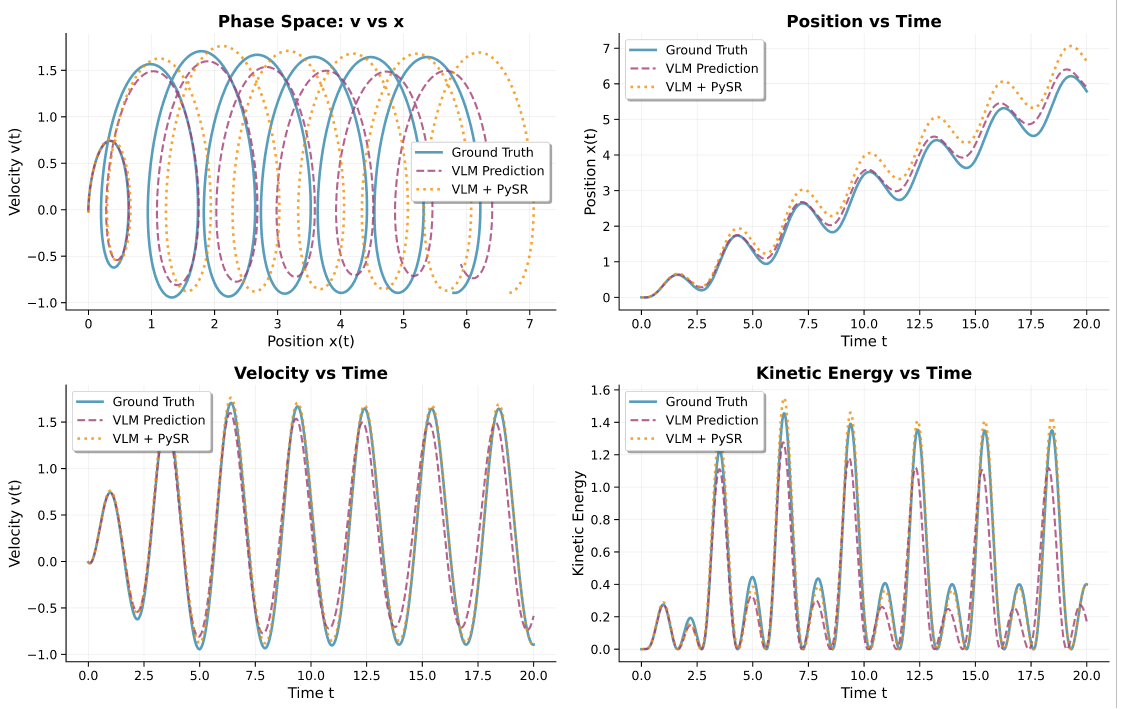}
    \caption{Case 3 System-Level Trajectory Comparison.}
    \label{fig:case_3_trajectory}
\end{figure}

\clearpage

\begin{figure}[h]
    \centering
    \includegraphics[width=\linewidth]{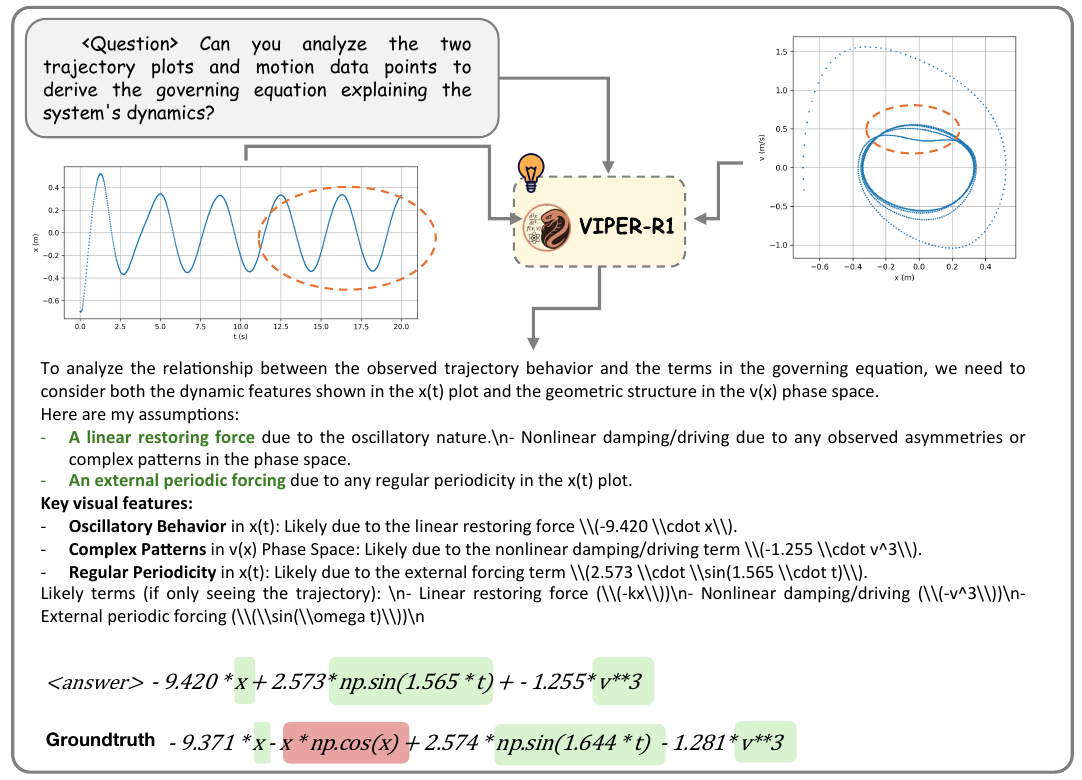}
    \caption{Case 4 Reasoning: A system with external periodic forcing.}
    \label{fig:case_4_reasoning}
\end{figure}

\begin{figure}[h]
    \centering
    \includegraphics[width=1\columnwidth]{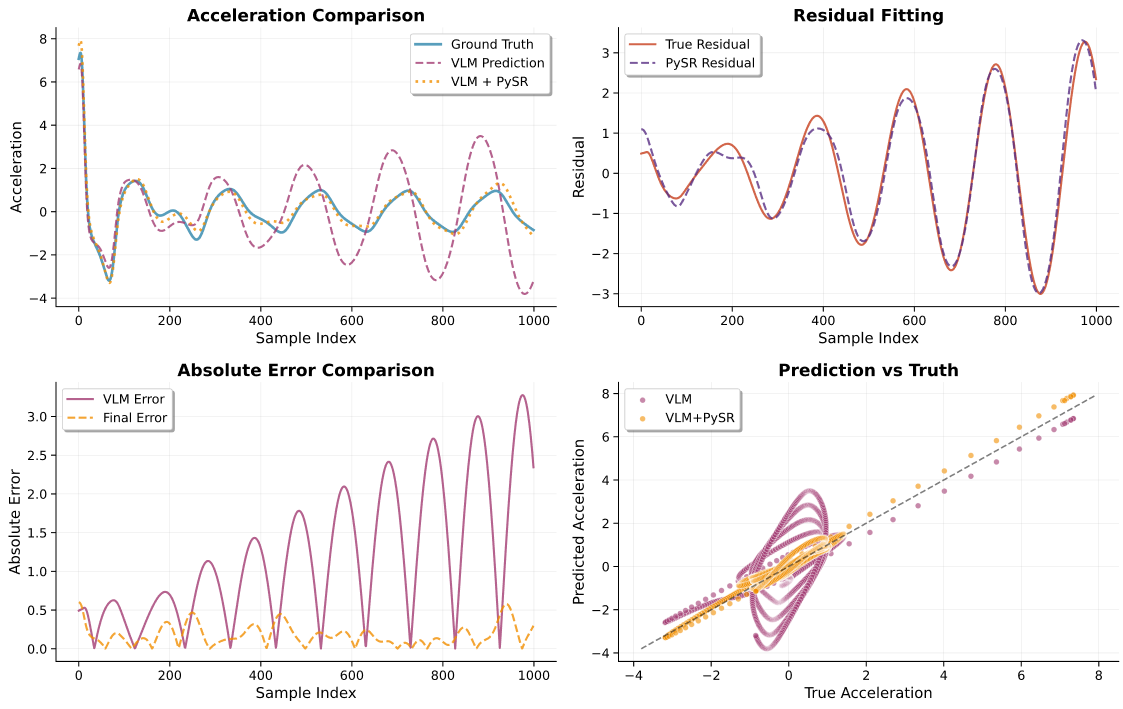}
    \caption{Case 4 Acceleration Signal Evaluation.}
    \label{fig:case_ana_4}
\end{figure}

\begin{figure}[h]
    \centering
    \includegraphics[width=1\columnwidth]{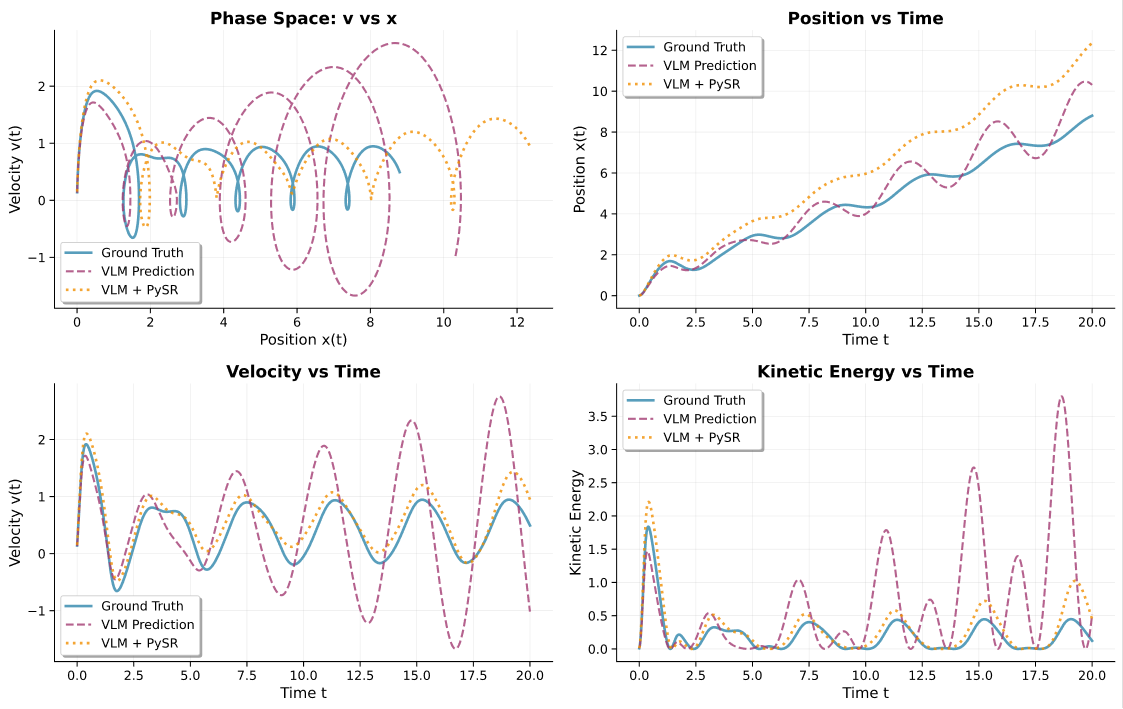}
    \caption{Case 4 System-Level Trajectory Comparison.}
    \label{fig:case_ana_4-2}
\end{figure}

\end{document}